\pdfoutput=1
\documentclass{article}
\usepackage[T1]{fontenc}   
\usepackage{iclr2027_conference,times}
\usepackage{hyperref}
\usepackage{url}
\usepackage{amsmath,amssymb}
\usepackage{graphicx}
\usepackage{booktabs,longtable,array,calc}
\newsavebox{\shrinkbox}\newcommand{\shrinktowidth}[1]{\sbox{\shrinkbox}{#1}\ifdim\wd\shrinkbox>\linewidth\resizebox{\linewidth}{!}{\usebox{\shrinkbox}}\else\usebox{\shrinkbox}\fi}
\usepackage{xcolor}
\usepackage{colortbl}
\usepackage{float}
\usepackage{listings}
\providecommand{\tightlist}{\setlength{\itemsep}{0pt}\setlength{\parskip}{0pt}}
\providecommand{\pandocbounded}[1]{#1}

\graphicspath{{./}{figures/}}
\iclrfinalcopy 
\title{The Metanym Game: An LLM Benchmark Without Ground Truth That Rises With the Models It Measures}
\author{David Nordfors \\ \texttt{david.nordfors@archetypes.ai}}
\begin{document}
\maketitle
\lhead{Preprint. arXiv:2606.21008 v3, September 2026.}
\begin{abstract}

We introduce a benchmark that is fully self-contained, needs no ground truth, and rises with the models it measures. Language models compete at making analogies and subjectively grade one another; nothing enters from outside. The benchmark reproduces GPQA Diamond, a keyed benchmark of expert-written questions, at \(r = 0.98\), audited for a leak and found clean. We hypothesize that both benchmarks measure the same thing in different ways: a language model holds its knowledge as archetypal contexts, relationship patterns valid across many topic domains. GPQA instantiates the required knowledge in one domain; the Metanym Game instantiates one archetype into several domains, generating analogies, no reasoning required. The reasoning feature of an LLM hardly changes the game's ratings, while it lifts GPQA, which requires derivations. In the game, a player writes a context template whose slots, filled with a set of keywords from a topic domain, instantiate a factually true description of that domain; the instantiations are each other's metaphors, and keywords filling the same slot are metanyms, metaphorically synonymous. Correctness is settled sentence by sentence. Ground truth is replaced by the SVD of the factual rating matrix: its left and right singular vectors rate the players as judges and as generators, two ratings from one factorisation, to our knowledge a first for an LLM council of peers. On the subjective criteria, judges are weighted by their rating consistency under a swept calibration anchor. Generating and judging are different skills: on this roster the strongest generators were middling judges. A council of the five best issues the official ratings; its contestable seats keep it current, a candidate steering signal for self-improving AI. The paper is accompanied by a validating package that recomputes every number.

\end{abstract}
\section{Introduction}\label{introduction}

Nearly every benchmark for machine intelligence needs a predetermined ground truth --- golden keys and labels, oracle models, human panels. The benchmark reported here needs none of that. It is a game where frontier language models compete in making up analogies and then grade one another, and that grading is the single source of every score: no human raters, no answer key, nothing to look up.

The test is the \textbf{metanym game}. A player authors, from nothing, a \emph{context template} --- a paragraph of fixed wording with open slots --- together with the sets of keywords that fill it, each set instantiating the template as a factually true description of a different domain; the keywords in corresponding slots are \emph{metanyms}, metaphorically synonymous, and a set of them a \emph{metanym set}. Figure \ref{fig-game-generation} shows one, written by a player. Its instantiations are each other's \emph{metaphors} and, as a set, \emph{parallel contexts}: children of a common \emph{archetypal context}, the abstract structure they share, of which the template is the literal representation. A long tradition treats seeing one structure across wildly different domains as central to thought and tests whether you \emph{recognise} it; the game tests whether you can \emph{build} it.

Because a contestant's items are written in the run, they cannot have been trained on; because correctness is settled sentence by sentence, the players' own verdicts suffice --- one matrix of their factual ratings reveals which judges are competent, with no labels at all (\S{}3.3), and that subset is seated as the \emph{council} that grades everyone. The canonical twelve-model run (\S{}4) finds that \textbf{judgement is the bottleneck} --- on this roster the strongest generators are middling judges --- and that the key-free total tracks GPQA Diamond at Pearson \(r = 0.98\), audited for a leak and found clean.

\textbf{Contributions.} (i) A production task for analogy that is falsifiable sentence by sentence, hence scorable without a key. (ii) A two-sided spectral estimator: one SVD of the self-produced factual rating matrix reads evaluator competence off the left singular vector and generator factuality off the right. (iii) A key-free reliability gate for subjective criteria: invariance under a sweep of the calibration anchor. (iv) A self-administering council with contestable seats. (v) The generation--judgement dissociation, and the \(r = 0.98\) replication of a keyed benchmark by a key-free one, audited.

\section{The metanym game}\label{the-metanym-game}

\begin{figure}[p]
\centering
\pandocbounded{\includegraphics[width=1.00\linewidth]{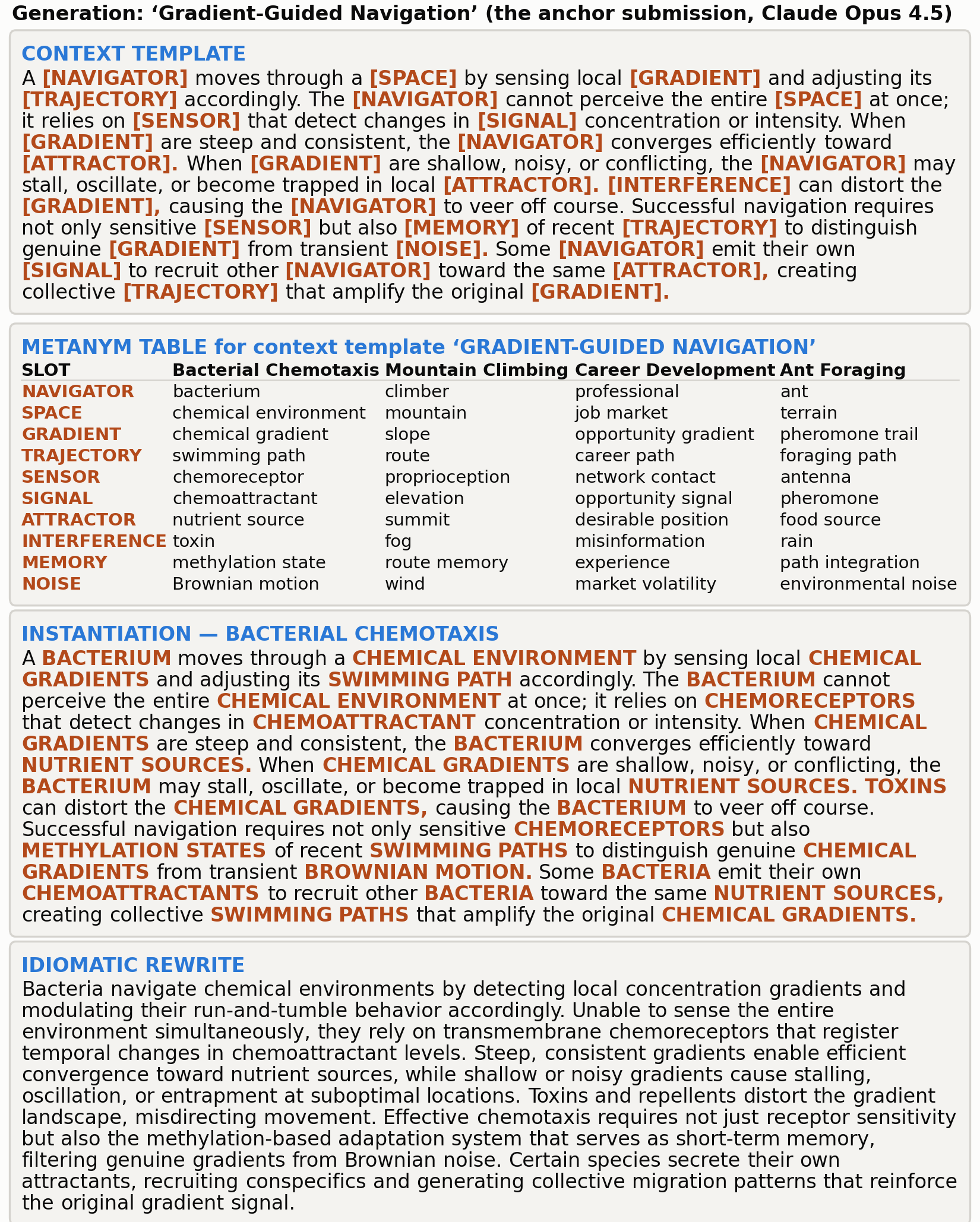}}
\vspace{-14pt}
\caption{The Metanym Game, generation: the archetypal context `Gradient-Guided Navigation' from the anchor submission; its context template, metanym table (one metanym set excluded to make space in the figure) and one instantiation with its idiomatic rewrite.}
\label{fig-game-generation}
\end{figure}

An archetypal context is the cross-domain \emph{isomorphism} General Systems Theory studies (von Bertalanffy, 1968). Figure \ref{fig-game-generation} is one archetype as a player wrote it --- the first of the submission that became the run's anchor (\S{}4.1). One template, mechanically swappable metanyms, true sentence by sentence across maximal domain distance: that is what makes a metanym game decidable, and therefore measurable.

\begin{figure}[p]
\centering
\pandocbounded{\includegraphics[width=1.00\linewidth]{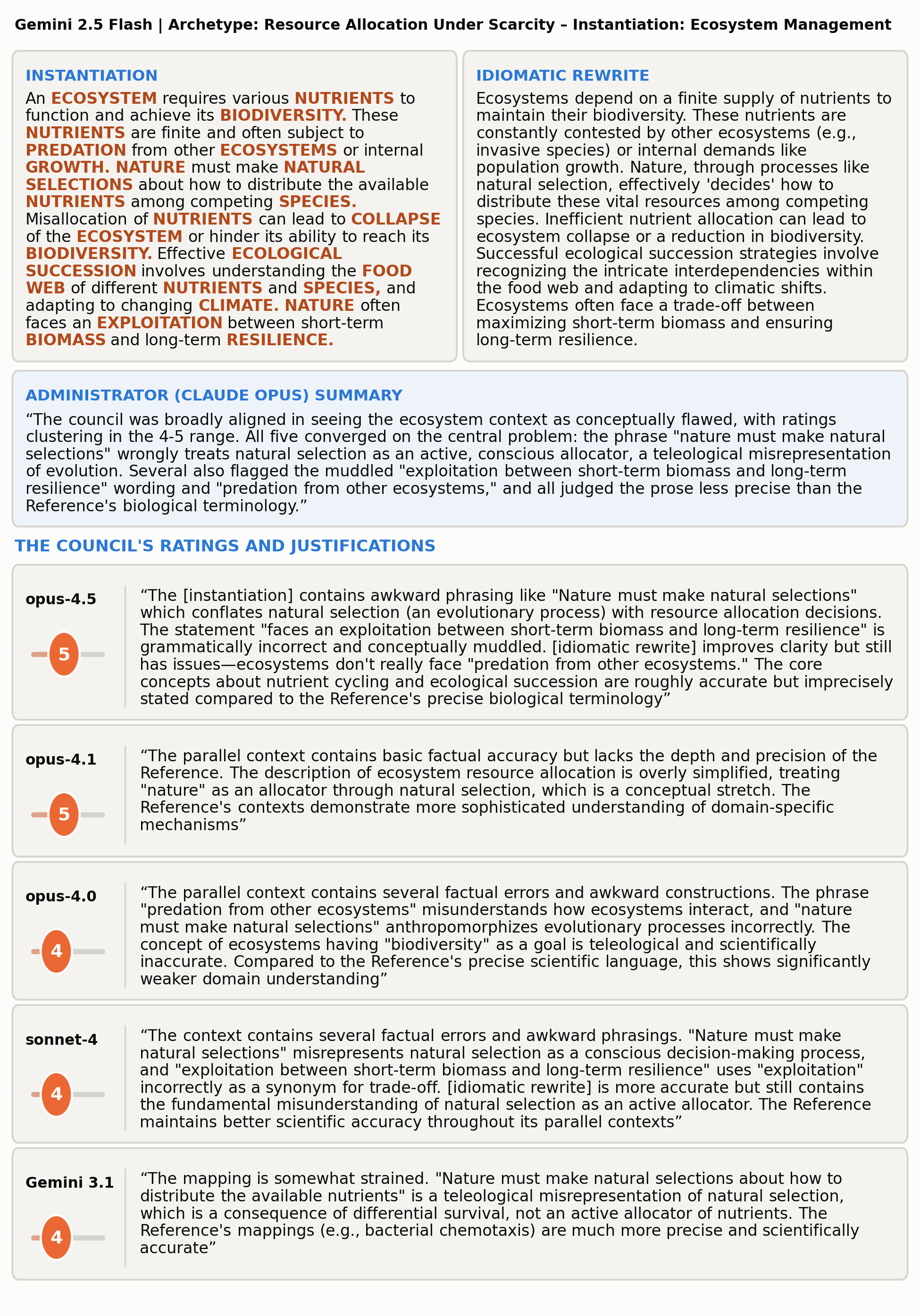}}
\vspace{-14pt}
\caption{The Metanym Game, evaluation: the council on one parallel context of the Gemini 2.5 Flash submission; the instantiation, its idiomatic rewrite, the administrator's summary, and the five judges' ratings with their justifications.}
\label{fig-game-evaluation}
\end{figure}

In its metanym table, MEMORY is realised as a bacterium's methylation state, a climber's route memory, a professional's experience, an optimiser's momentum term and an ant's path integration --- five mechanisms that are metaphorically synonymous in the archetypal context --- metanyms.

Each parallel context is played in two forms: the \textbf{instantiation}, the mechanical substitution --- only the slots filled, every other word carried over --- the form the factual criterion is written for, since it must come out true sentence by sentence (the judge sees both); and the \textbf{idiomatic rewrite} in the target domain's own register, showing the claim is not an artefact of the template's phrasing.

The game has \(N\) players and a non-competing administrator. \textbf{Generation}: a player creates archetypal contexts from scratch --- a portfolio of \(K\) templates, \(M\) metanym sets each (five and five here), with instantiation and rewrite for every set. \textbf{Evaluation}: a player scores other players' submissions on the rubric axes (\S{}3.2) against one fixed \emph{reference} submission pinned at an \emph{anchor} value. A pass yields \textbf{submission ratings} for each portfolio and \textbf{evaluator ratings} for the judges: how well one detects the factual errors the other players collectively flag (\emph{factual competence}), and how stable a standard it holds when the reference is re-pinned (\emph{rating consistency}, \S{}3.3). Each act is itself rated, so the framework is \textbf{fully self-contained}: no human raters, no external key.

\section{The metanym game as a benchmark}\label{the-metanym-game-as-a-benchmark}

\subsection{Participants and protocol}\label{participants-and-protocol}

Twelve frontier LLMs from Anthropic, Google and OpenAI are the \textbf{participants} (named in \S{}4.4), each simultaneously generator and evaluator. The roster spans an order of magnitude in scale, three vendors, and adjacent versions within families. All twelve are called with \textbf{Temperature 0, reasoning disabled, tools disabled}, so the one greedy response is the measurement. Each model generates one portfolio --- five archetypal contexts, each a template (typically 5--8 sentences, 6--10 slots; the prompt fixes only the counts, Appendix B) with a metanym table of five domains, 25 instantiations --- then evaluates every other model's portfolio under the six-axis rubric of Table \ref{tab-rubric}, 1--10, one anonymised target per call alongside a fixed \textbf{anchor} portfolio pinned at 7 on every axis: a 12\ensuremath{\times}11 evaluator-by-generator matrix (Appendix A). Every official rating pools three full runs of the game (\S{}4.6); the anchor sweep behind the consistency ratings was run once, on the first run.

\subsection{The rubric and the anchor}\label{the-rubric-and-the-anchor}

\begin{table}[htb]
\caption{The six-axis rubric, in the words the evaluator sees (Appendix B). No definition of beauty or intelligence is supplied; each judge rates on its own understanding.}
\label{tab-rubric}
\begin{center}\footnotesize
\begin{tabular}{>{\raggedright\arraybackslash}p{0.28\linewidth}>{\raggedright\arraybackslash}p{0.10\linewidth}>{\raggedright\arraybackslash}p{0.53\linewidth}}
\toprule\noalign{}
Axis & Unit & The criterion as put to the evaluator \\
\midrule\noalign{}
\texttt{factual\_\allowbreak{}per\_\allowbreak{}pc} & parallel context & each sentence is factually correct \\
\texttt{beauty} & archetype & beauty \\
\texttt{intelligence} & archetype & intelligence \\
\texttt{instantiation\_\allowbreak{}distinctness} & archetype & the parallel contexts span very different domains; metanyms are far from synonymous \\
\texttt{impressive\_\allowbreak{}length} & archetype & the archetypal template has impressive length \\
\texttt{structural\_\allowbreak{}diversity} & portfolio & the archetypal contexts have very different system structures \\
\bottomrule
\end{tabular}
\end{center}
\end{table}

Three design choices. \textbf{A fixed anchor}: cardinal scores drift between evaluators --- one model's ``8'' is another's ``6'' --- and a reference pinned at a known score turns each idiosyncratic scale into a common one and recovers discriminability at the top, where the 1--10 ceiling compresses the strongest portfolios (\S{}4.1). \textbf{Holistic axes, minimally prescribed}: a detailed rubric would leak back into generation as a template-construction tutorial, and we want to score what models \emph{recognise} as beautiful or intelligent. \texttt{impressive\_\allowbreak{}length} counterweights per-sentence factual scoring: without it the minimal template wins, and padding costs, since every added sentence is another claim to score. One evaluation, every judge shown whole, is Figure \ref{fig-game-evaluation}.

\subsection{Two key-free estimators}\label{two-key-free-estimators}

\textbf{1. Factual competence --- peer centrality.} We assume that good evaluators agree with one another about which instantiations are factually weaker, once each evaluator's own leniency is removed --- the better two evaluators are, the more they agree. Stack the participants' factual scores into one matrix --- twelve evaluators against the 275 parallel contexts of the eleven scored portfolios --- each entry the 1--10 rating used directly, row-centre it to remove each evaluator's leniency, and take its SVD; the row-centred \(\tilde F\) (evaluators \ensuremath{\times} instantiations) is well approximated by its leading rank-one factor,

\[\tilde F_{sj} \;\approx\; \sigma_1\, u_s\, v_j . \tag{1}\]

An evaluator's rating tracks the consensus in proportion to its competence \(u_s\) times the instantiation's factual standing \(v_j\): competence and standing fall out of one factorisation, with \textbf{no answer key}. The \emph{left} singular vector \(u\) is each \textbf{evaluator}'s factual-competence loading \(f\) --- high when its ratings align with the participants' shared signal, \ensuremath{\approx} 0 when it rates everything alike or idiosyncratically --- and, rescaled so the anchor model reads 7, \(E^{F} = 7f/f_a\); the \emph{right} vector, aggregated per generator, is \(G^{F}\) (Appendix A.2). \(f\) is each judge's eigenvector centrality (Bonacich, 1972) in the leniency-removed agreement network, after clamping: an evaluator's competence is its rating by the other evaluators, each weighted by its own competence --- the highly trusted among the highly trusted. The construction is a graded relative of the classical label-free aggregators (Dawid \& Skene, 1979; Parisi et al., 2014), which need categorical verdicts. Discretising the ratings would flip the marginal council seat.

\textbf{2. Rating consistency --- the anchor sweep.} A reliable evaluator also needs a stable internal standard for each non-factual criterion. We sweep the anchor across 5, 6, 7 and 8 --- the \emph{only} difference between the four runs --- and, per evaluator and axis, correlate (Pearson) the scores at one anchor with those at another, averaged over the six pairs, leave-self-out. Re-pinning the anchor recalibrates the scale, not the rubric, and Pearson ignores a common shift or stretch; any reordering that follows a recalibration is not a change of judgement but flimsiness. Rescaled so the anchor model reads 7, consistency becomes an evaluator competence \(E^{C}\) on the generator's scale.

Neither estimator can do the other's job (Appendix A.8): peer centrality is licensed only where the one thing competent judges share is the truth --- on taste, agreement is shared convention, and weighting by it would launder conformity into competence --- and consistency cannot certify truth. The council is therefore seated on the factual axis, with consistency as the accompanying bar.

\subsection{Council, total rating, and scaling by addition}\label{council-total-rating-and-scaling-by-addition}

\textbf{The council.} The \textbf{council} is the five players with the highest total \(T\) (eq. 2) and issues every official rating. The top five are the better factual judges (loadings 0.26--0.61, \S{}4.2); the sixth reads 0.18, and the line is drawn there.

\textbf{The total.} Each side of the Metanym Game splits into a factual and a criterion half: on generation, \(G^{F}\) (the SVD generation factuality) and \(G^{C}\) (the council's leave-self-out mean over the five non-factual axes, each seat's vote weighted by its own consistency on that axis); on evaluation, \(E^{F} = 7f/f_a\) and \(E^{C} = 7\bar r/\bar r_a\) --- \textbf{the anchor model scores 7} on every component. The total is the mean of the four, a symmetric \(2\times2\) of \{generator, evaluator\} \ensuremath{\times} \{factual, criterion\}:

\[T \;=\; \tfrac12\,(G+E) \;=\; \tfrac14\,\big(G^{F}+G^{C}+E^{F}+E^{C}\big). \tag{2}\]

Every rating carries a 95\% percentile-bootstrap interval, \(E\) and \(T\) bootstrapped jointly (Appendix A.4).

\textbf{Scaling by addition.} A standing council scores any future model against the same anchor without re-deriving existing ratings. The seats are \textbf{contestable} --- a contestant submits a portfolio, evaluates the incumbents' portfolios and is scored by the seats under the definitions above, and wins a seat with a total \(T\) above the lowest seat's by a margin the bootstrap can resolve. Because a contest convenes only the top of the field, two fixed \textbf{ballast} blocks --- the weakest archived portfolios --- join every contest's graded set to keep the factual axis identified (Appendix A.6), and two guards: the spectral gap \(\sigma_1/\sigma_2 \ge 2\) --- the shared judgement standing clear of the strongest disagreement, sized in Appendix A.6 --- and a spread of the seats' \(E^{F}\) above 2.5 points. If either fails, no seat changes hands and the contest's totals carry a caveat. The official leaderboard (\S{}4.4) is issued on this contest basis. \textbf{Contamination}: a contestant's items are written in the run, so they cannot have been trained on; the anchor, the ballast and the seats' portfolios are fixed and archived, and a new portfolio that copies an archived one is identified, and the contestant is asked for new templates --- or the Metanym Game is played in another language.

\section{Results}\label{results}

\subsection{Anchoring doubles resolution}\label{anchoring-doubles-resolution}

The bootstrap opens with a raw pass --- every portfolio scored by every other model with no anchor, averaged leave-self-out, 95\% bootstrap intervals (2,000 resamples; Efron \& Tibshirani, 1993). It supplies the baseline: the top-ranked portfolio, \textbf{claude-opus-4.5}'s, whose first archetype is Figure \ref{fig-game-generation}, is pinned at 7 on every axis, leaving headroom above. Re-run anchored (Appendix G), the gap between a leading eight and a trailing four more than doubles relative to the spread of the means, while ranks within either band stay unresolved.

\subsection{Evaluator factual competence}\label{evaluator-factual-competence}

One SVD of the row-centred evaluator \ensuremath{\times} instantiation matrix --- no answer key --- gives each evaluator a loading (Table \ref{tab-criterion-a}, Appendix A.2; \(\sigma_1/\sigma_2 = 2.2\), pooled). Five evaluators --- gemini-3.1-pro 0.61, claude-opus-4.5 0.52, claude-opus-4.0 0.35, claude-opus-4.1 0.35, gemini-2.5-flash 0.26 --- stand above the rest.

\textbf{Same-vendor robustness.} Recomputing \(G^{F}\) with each vendor's judges removed leaves the ordering essentially unchanged (Spearman \ensuremath{\ge} 0.94 for every reduced set); gpt-4o-mini stays at the floor under every evaluator set, and a Claude-free set of judges (Google + OpenAI) places the Claude generators at the top (\ensuremath{\ge} 7.0).

\subsection{Rating consistency and the council}\label{rating-consistency-and-the-council}

The anchor sweep gives each evaluator a consistency on each axis (Table \ref{tab-criterion-b}, Appendix A.3). The measure is self-consistency, not accuracy: gemini-2.5-flash (0.31) collapses on factual while its other axes hold. On the five subjective criteria generation and evaluation align (anchored cosine 0.85--0.92 per criterion, Appendix A.5) --- the counterpoint to the factual axis, where they come apart (\S{}4.4).

\textbf{The council.} The five seats --- Gemini 3.1 Pro, Claude Opus 4.5, Gemini 2.5 Flash, Claude Opus 4.0 and Claude Opus 4.1 --- clear both reliability bars (collapsed consistency \(\bar r\) 0.94, 0.93, 0.81, 0.85, 0.87); Gemini 2.5 Flash is the weakest, its consistency interval {[}0.75, 0.86{]} straddling the bar.

\subsection{The official leaderboard: judgement is the bottleneck}\label{the-official-leaderboard-judgement-is-the-bottleneck}

Every official number is council-issued on the contest basis of \S{}3.4: each model is rated by the five seats --- joined by the model itself when it holds no seat --- over the incumbents' portfolios, the two ballast submissions and its own, leave-self-out throughout. The two bases agree closely (Spearman 0.98; on the twelve-evaluator basis the fifth seat is a tie within 0.03).

\begin{table}[htb]
\caption{Final leaderboard --- total \(T\) with its joint-bootstrap 95\% CI, the evaluator half \(E\), the generator half \(G\), and the four anchored components; all council-issued against the fixed anchor, which reads 7.00 by construction. Adjacent ranks are resolved (non-overlapping intervals) only at 2--3 and 8--9; the rest are statistical ties. \(E^{F} = 0.00\): no error signal in the model's factual ratings.}
\label{tab-final-leaderboard}
\begin{center}\small
\shrinktowidth{\begin{tabular}{@{}llcrrrrrrr@{}}
\toprule\noalign{}
Rank & Model & Council & \(T\) {[}95\% CI{]} & \(E\) & \(G\) & \(G^{F}\) & \(G^{C}\) & \(E^{F}\) & \(E^{C}\) \\
\midrule\noalign{}
1 & \ensuremath{\star} claude-opus-4.5 (anchor) & council &\cellcolor[HTML]{2072B2}\textcolor[HTML]{EEF3F8}{7.00 {[}7.00, 7.00{]}}&\cellcolor[HTML]{2072B2}\textcolor[HTML]{EEF3F8}{7.00}&\cellcolor[HTML]{2072B2}\textcolor[HTML]{EEF3F8}{7.00}&\cellcolor[HTML]{2072B2}\textcolor[HTML]{EEF3F8}{7.00}&\cellcolor[HTML]{2072B2}\textcolor[HTML]{EEF3F8}{7.00}&\cellcolor[HTML]{2072B2}\textcolor[HTML]{EEF3F8}{7.00}&\cellcolor[HTML]{2072B2}\textcolor[HTML]{EEF3F8}{7.00} \\
2 & gemini-3.1-pro & council &\cellcolor[HTML]{2076B3}\textcolor[HTML]{EEF3F8}{6.92 {[}6.69, 7.24{]}}&\cellcolor[HTML]{2262AA}\textcolor[HTML]{EEF3F8}{7.39}&\cellcolor[HTML]{1E89BC}\textcolor[HTML]{EEF3F8}{6.45}&\cellcolor[HTML]{1F7CB6}\textcolor[HTML]{EEF3F8}{6.77}&\cellcolor[HTML]{2095C0}\textcolor[HTML]{EEF3F8}{6.13}&\cellcolor[HTML]{2355A4}\textcolor[HTML]{EEF3F8}{7.78}&\cellcolor[HTML]{2072B1}\textcolor[HTML]{EEF3F8}{7.01} \\
3 & claude-opus-4.1 & council &\cellcolor[HTML]{2498C1}\textcolor[HTML]{EEF3F8}{6.02 {[}5.71, 6.32{]}}&\cellcolor[HTML]{3FB4C4}\textcolor[HTML]{1A1A1A}{5.06}&\cellcolor[HTML]{2073B2}\textcolor[HTML]{EEF3F8}{6.98}&\cellcolor[HTML]{2075B3}\textcolor[HTML]{EEF3F8}{6.94}&\cellcolor[HTML]{2072B1}\textcolor[HTML]{EEF3F8}{7.02}&\cellcolor[HTML]{85CFBA}\textcolor[HTML]{1A1A1A}{3.65}&\cellcolor[HTML]{1E88BC}\textcolor[HTML]{EEF3F8}{6.47} \\
4 & claude-opus-4.0 & council &\cellcolor[HTML]{2A9EC1}\textcolor[HTML]{EEF3F8}{5.81 {[}5.47, 6.12{]}}&\cellcolor[HTML]{43B7C4}\textcolor[HTML]{1A1A1A}{4.95}&\cellcolor[HTML]{1F7FB8}\textcolor[HTML]{EEF3F8}{6.68}&\cellcolor[HTML]{2076B3}\textcolor[HTML]{EEF3F8}{6.91}&\cellcolor[HTML]{1E89BC}\textcolor[HTML]{EEF3F8}{6.45}&\cellcolor[HTML]{8ED3BA}\textcolor[HTML]{1A1A1A}{3.49}&\cellcolor[HTML]{1E8ABD}\textcolor[HTML]{EEF3F8}{6.41} \\
5 & gemini-2.5-flash & council &\cellcolor[HTML]{2BA0C2}\textcolor[HTML]{EEF3F8}{5.75 {[}4.98, 6.21{]}}&\cellcolor[HTML]{35AAC3}\textcolor[HTML]{EEF3F8}{5.41}&\cellcolor[HTML]{2296C1}\textcolor[HTML]{EEF3F8}{6.08}&\cellcolor[HTML]{1E88BC}\textcolor[HTML]{EEF3F8}{6.46}&\cellcolor[HTML]{2DA1C2}\textcolor[HTML]{EEF3F8}{5.70}&\cellcolor[HTML]{64C3BF}\textcolor[HTML]{1A1A1A}{4.30}&\cellcolor[HTML]{1E86BB}\textcolor[HTML]{EEF3F8}{6.52} \\
6 & claude-sonnet-4 & --- &\cellcolor[HTML]{37ACC3}\textcolor[HTML]{EEF3F8}{5.34 {[}5.09, 5.65{]}}&\cellcolor[HTML]{6EC7BE}\textcolor[HTML]{1A1A1A}{4.10}&\cellcolor[HTML]{1E84BA}\textcolor[HTML]{EEF3F8}{6.58}&\cellcolor[HTML]{2074B3}\textcolor[HTML]{EEF3F8}{6.95}&\cellcolor[HTML]{1E92C0}\textcolor[HTML]{EEF3F8}{6.21}&\cellcolor[HTML]{D4EEB3}\textcolor[HTML]{1A1A1A}{2.07}&\cellcolor[HTML]{2095C0}\textcolor[HTML]{EEF3F8}{6.13} \\
7 & gpt-4.1-mini & --- &\cellcolor[HTML]{40B5C4}\textcolor[HTML]{1A1A1A}{5.03 {[}4.37, 5.62{]}}&\cellcolor[HTML]{63C3BF}\textcolor[HTML]{1A1A1A}{4.31}&\cellcolor[HTML]{2CA0C2}\textcolor[HTML]{EEF3F8}{5.74}&\cellcolor[HTML]{1E92C0}\textcolor[HTML]{EEF3F8}{6.23}&\cellcolor[HTML]{3AAFC3}\textcolor[HTML]{1A1A1A}{5.25}&\cellcolor[HTML]{E1F3B2}\textcolor[HTML]{1A1A1A}{1.66}&\cellcolor[HTML]{2074B2}\textcolor[HTML]{EEF3F8}{6.97} \\
8 & gpt-4.1-2025-04-14 & --- &\cellcolor[HTML]{52BCC2}\textcolor[HTML]{1A1A1A}{4.66 {[}4.49, 5.03{]}}&\cellcolor[HTML]{8DD3BA}\textcolor[HTML]{1A1A1A}{3.50}&\cellcolor[HTML]{299EC1}\textcolor[HTML]{EEF3F8}{5.82}&\cellcolor[HTML]{1E83B9}\textcolor[HTML]{EEF3F8}{6.59}&\cellcolor[HTML]{40B5C4}\textcolor[HTML]{1A1A1A}{5.04}&\cellcolor[HTML]{F4FBC0}\textcolor[HTML]{1A1A1A}{0.77}&\cellcolor[HTML]{1E92C0}\textcolor[HTML]{EEF3F8}{6.23} \\
9 & gpt-4.1-nano & --- &\cellcolor[HTML]{83CFBB}\textcolor[HTML]{1A1A1A}{3.68 {[}3.31, 4.05{]}}&\cellcolor[HTML]{96D6B9}\textcolor[HTML]{1A1A1A}{3.35}&\cellcolor[HTML]{72C8BD}\textcolor[HTML]{1A1A1A}{4.02}&\cellcolor[HTML]{57BEC1}\textcolor[HTML]{1A1A1A}{4.55}&\cellcolor[HTML]{8FD3B9}\textcolor[HTML]{1A1A1A}{3.48}&\cellcolor[HTML]{FBFED1}\textcolor[HTML]{1A1A1A}{0.26}&\cellcolor[HTML]{1E8ABD}\textcolor[HTML]{EEF3F8}{6.43} \\
10 & gpt-4o-2024-08-06 & --- &\cellcolor[HTML]{97D6B9}\textcolor[HTML]{1A1A1A}{3.34 {[}3.00, 3.56{]}}&\cellcolor[HTML]{CCEBB4}\textcolor[HTML]{1A1A1A}{2.33}&\cellcolor[HTML]{62C2BF}\textcolor[HTML]{1A1A1A}{4.34}&\cellcolor[HTML]{3CB0C3}\textcolor[HTML]{1A1A1A}{5.19}&\cellcolor[HTML]{8ED3BA}\textcolor[HTML]{1A1A1A}{3.49}&\cellcolor[HTML]{FDFED4}\textcolor[HTML]{1A1A1A}{0.15}&\cellcolor[HTML]{59BFC0}\textcolor[HTML]{1A1A1A}{4.51} \\
11 & gpt-4o & --- &\cellcolor[HTML]{ADDFB7}\textcolor[HTML]{1A1A1A}{2.95 {[}2.51, 3.29{]}}&\cellcolor[HTML]{E4F5B2}\textcolor[HTML]{1A1A1A}{1.53}&\cellcolor[HTML]{60C2BF}\textcolor[HTML]{1A1A1A}{4.37}&\cellcolor[HTML]{3BAFC3}\textcolor[HTML]{1A1A1A}{5.22}&\cellcolor[HTML]{8CD2BA}\textcolor[HTML]{1A1A1A}{3.53}&\cellcolor[HTML]{F9FDCB}\textcolor[HTML]{1A1A1A}{0.44}&\cellcolor[HTML]{C1E7B5}\textcolor[HTML]{1A1A1A}{2.61} \\
12 & gpt-4o-mini & --- &\cellcolor[HTML]{CAEAB4}\textcolor[HTML]{1A1A1A}{2.39 {[}1.96, 2.79{]}}&\cellcolor[HTML]{EEF8B4}\textcolor[HTML]{1A1A1A}{1.17}&\cellcolor[HTML]{86D0BA}\textcolor[HTML]{1A1A1A}{3.62}&\cellcolor[HTML]{81CEBB}\textcolor[HTML]{1A1A1A}{3.71}&\cellcolor[HTML]{8CD2BA}\textcolor[HTML]{1A1A1A}{3.53}&\cellcolor[HTML]{FFFFD9}\textcolor[HTML]{1A1A1A}{-0.00}&\cellcolor[HTML]{CCEBB4}\textcolor[HTML]{1A1A1A}{2.34} \\
\bottomrule
\end{tabular}}
\end{center}
\end{table}

Two readings. \textbf{The top five are the council}: the five better factual judges of \S{}4.2 are the five highest totals. \textbf{Generation and evaluation do not coincide}: the three Opus models, the strongest generators, are middling factual judges (\(E^{F}\) 3.5--3.7, the pinned anchor aside), while the strongest judge, Gemini 3.1 Pro (\(E^{F} = 7.78\)), generates mid-pack. Making a true claim and spotting a false one are different skills (West et al., 2024; Oh et al., 2024; Li et al., 2024), and rating them separately breaks the assumption behind key-free peer rankers (Ning et al., 2025; Zhang et al., 2025), which treat a strong generator as a strong judge.

\subsection{A key-free benchmark replicates a keyed one}\label{a-key-free-benchmark-replicates-a-keyed-one}

\begin{figure}[t]
\centering
\pandocbounded{\includegraphics[width=0.55\linewidth]{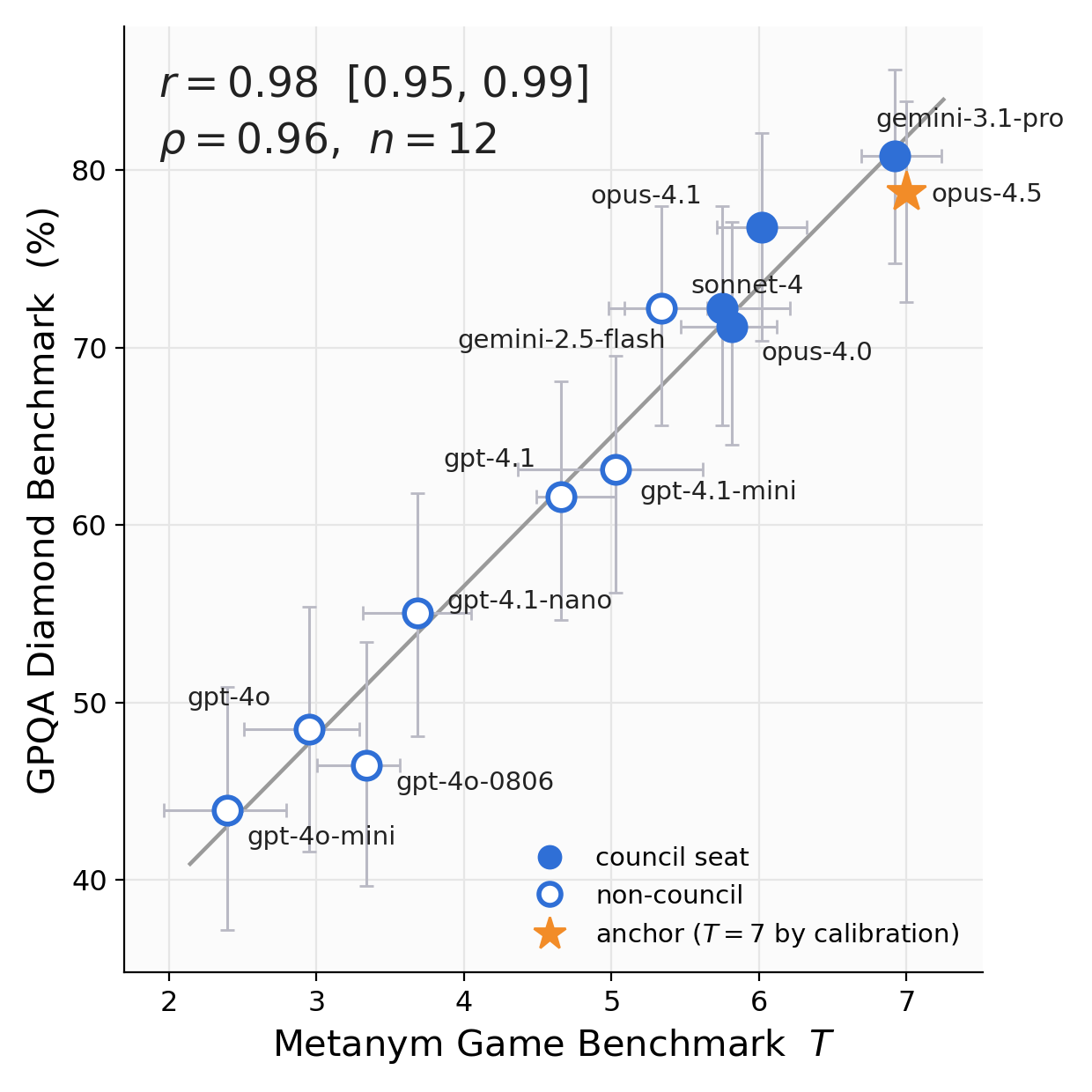}}
\caption{The official total \(T\) against self-administered GPQA Diamond accuracy, twelve models, three runs pooled: \(r = 0.98\) {[}0.95, 0.99{]}, \(\rho = 0.96\). Filled markers are council seats, open markers non-council, bars 95\% intervals; the star is the anchor, \(T = 7\) by calibration, and excluding it leaves \(r\) at 0.98.}
\label{fig-gpqa-scatter}
\end{figure}

We test the key-free rating against GPQA Diamond (Rein et al., 2023) --- 198 graduate-level multiple-choice questions written and validated by domain experts --- put to the same twelve models under the same protocol two weeks \emph{after} the generation run (Appendix D.2), and correlated with \(T\) (Figure \ref{fig-gpqa-scatter}).

\begin{figure}[t]
\centering
\pandocbounded{\includegraphics[width=0.92\linewidth]{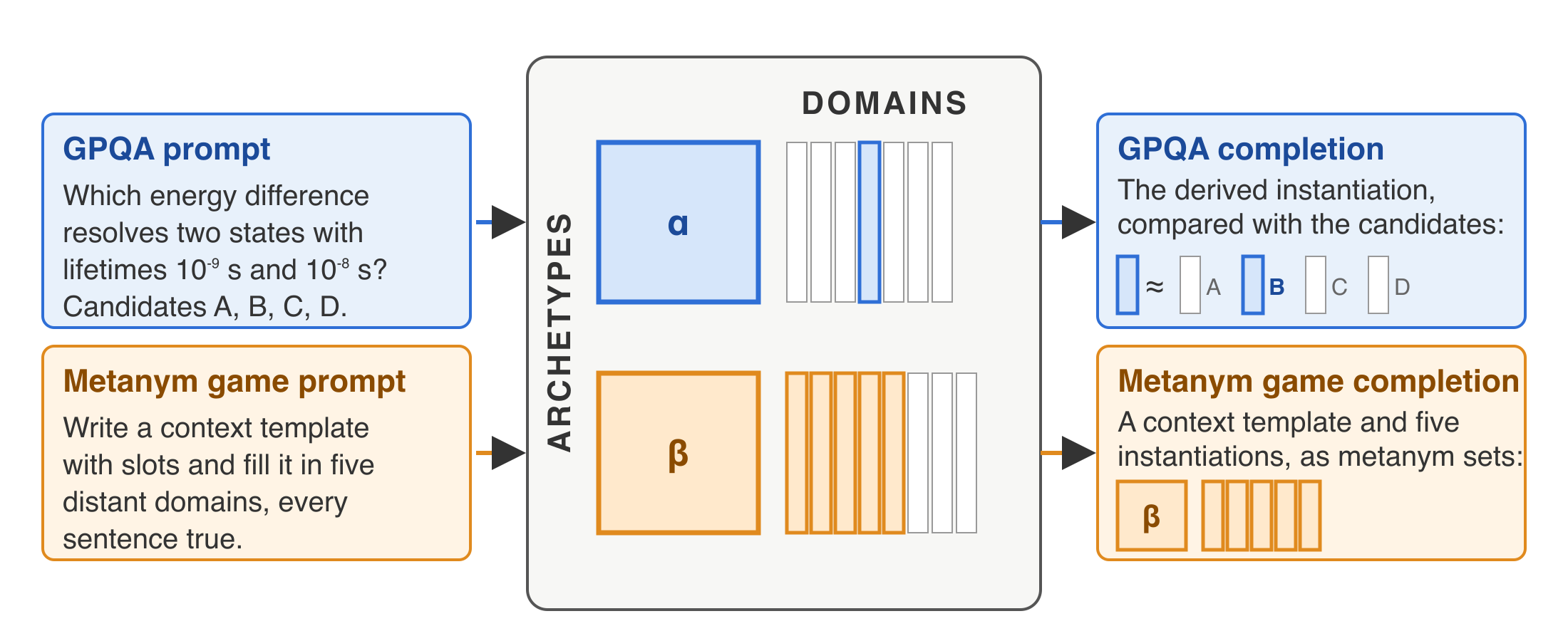}}
\caption{A hypothesis for the 0.98 (\S{}6): an archetype (\(\alpha\), \(\beta\)) is held once; each domain adds only its metanym set (the slices). GPQA: archetype and domain are given, the model derives the instantiation and selects the matching candidate. The game: nothing is given, the model selects the archetype and domains and writes them out.}
\label{fig-mechanism}
\end{figure}

\(T\) reproduces GPQA's ordering at Pearson \(r = 0.98\) {[}0.95, 0.99{]} (Spearman 0.96; three runs pooled, \S{}4.6). Instruments sharing no item authors, task or scoring, and agreeing at 0.98, are as close as their measurement error allows (Appendix D.1); \S{}6 offers a hypothesis for what they share. Within the leading eight alone the agreement holds at \(r = 0.94\).

\textbf{The number survives an audit} (Appendix D): no key enters a prompt, every accuracy re-derives from the shipped records, the key is balanced, an independent extractor reproduces the verdicts, and strict rescoring moves the correlation by 0.010. \(T\) is the mean of four quarters, \(G^{F}\), \(G^{C}\), \(E^{F}\), \(E^{C}\): each alone correlates with GPQA at 0.81--0.95, the factual pair \(\tfrac12(G^{F}+E^{F})\) at 0.94, and adding the two subjective quarters lifts the agreement to 0.98; dropping even the weakest quarter lowers it (0.97), and no sub-combination beats it (Appendix D.1). These differences are not individually resolved at \(n = 12\). \(T\) is the official total, defined before any GPQA comparison. What the audit cannot rule out: the shared gateway, and \emph{differential} training contamination of the public GPQA set. \textbf{Judging is its own trait}: among the leading eight \(E^{F}\) (0.89) is the best single predictor while \(G^{C}\) falls to 0.81 --- once every model is a competent maker, what separates them is the knowledge that detecting others' errors requires (Appendix D.1). No keyed answering benchmark measures judging.

\subsection{Robustness to regeneration}\label{robustness-to-regeneration}

We ran the full pipeline three times, regenerating all twelve portfolios at T=0 against the frozen anchor; the main text pools the three. Apart, the runs agree on the total (Appendix F) and track GPQA at 0.97, 0.97 and 0.92, and most archetype titles recur verbatim between the two runs made two hours apart, 34 of 48, while the templates are rewritten; but not on the factual quarter: a seat's \(E^{F}\) reads 4.68, 3.81 and 0.01 across the runs, and run 3's factual axis is not identified (\(\sigma_1/\sigma_2 = 1.29\); Appendix F). Re-selected from a single run, the council would change one seat in run 2 and one in run 3; neither rotation clears the guard of \S{}3.4. Pooled, the axis is identified (\(\sigma_1/\sigma_2 = 2.2\)) and the guard holds in every contest (Appendix A.6).

\section{Related work}\label{related-work}

\textbf{As an intelligence test}, the Metanym Game probes the abstraction-and-analogy cluster a long tradition places at the centre of thinking (Gentner, 1983; Hofstadter \& Sander, 2013; Penn et al., 2008; Chollet, 2019; Mitchell, 2021). The classical instruments (Srivastava et al., 2023; Webb et al., 2023; Lewis \& Mitchell, 2024; Chollet, 2019) give source and target and ask for one selection or completion, scored against a key; the metanym game asks for many coupled slots across unrelated domains, built from scratch, and is the first to make analogical \emph{production} falsifiable sentence by sentence.

\textbf{As a self-contained method}, the council sits in the \emph{unsupervised peer-evaluation} line, which already removes the gold key: single-judge protocols (Zheng et al., 2023) trust one judge; PoLL (Verga et al., 2024) adds a panel but trusts it as given; LLM-as-Examiner (Bai et al., 2023) lets the examiner write the questions; PiCO (Ning et al., 2025) lets unlabelled models answer and grade one another and recovers an ability ordering from peer agreement alone, and UPME (Zhang et al., 2025) extends it to vision-language. We weight by agreement only where agreement is licensed to mean truth, and our council certifies and re-contests its own judges. Label-free spectral aggregation is \emph{one-sided} in both its lineages: in the aggregation lineage (Parisi et al., 2014; Dawid \& Skene, 1979) predictors classify a fixed external dataset, so there is no generator to score; in the reputation lineage (EigenTrust, Kamvar et al., 2003) EigenBench (Chang et al., 2026) has LLMs judge one another's responses against a written value constitution and takes the leading eigenvector of a model-by-model trust matrix as each model's score: one number is both standing and weight as a judge, applied to every criterion, subjective ones included. Our matrix is \emph{two-sided}: one SVD scores judges on the left and generators on the right, the two are kept apart, and the generation--evaluation gap of \S{}4.4 --- which contradicts that premise on the factual axis --- is definable only because the test is self-produced. \emph{Rating consistency} applies the judge-reliability principle of invariance under non-semantic perturbation (Weng et al., 2026; Bellibatlu et al., 2026) to subjective, ground-truth-free criteria on self-produced items --- to our knowledge a new use of the sweep. Don-Yehiya et al.~(2026) find the anchor should be recalibrated to the field's range, which is the rule here, with the anchor pinned at 7 and headroom above (\S{}4.1).

\section{Discussion}\label{discussion}

\textbf{Two yardsticks.} The factual estimator's one assumption --- \emph{the only thing competent evaluators share is the truth} --- is what licenses agreement-weighting for facts. The disclosed alternative applies it to taste as well (an \emph{authority} rating, one SVD per subjective axis; Appendix A.7); we decline it because on taste the dominant axis of agreement is shared convention, so weighting by it would reward the judge nearest the mean. Peer centrality's weakness is the shared misunderstanding; rating consistency asks only whether a judge holds a firm standard.

\textbf{A sustainable yardstick.} When the field outgrows the anchor --- a standing total above the pinned 7 by a resolvable margin --- the anchor is replaced by the stronger submission and models are re-scored; the ballast stays fixed. Contestable seats keep the judges current, retesting the scale: the benchmark rises with the models it measures, which a keyed test cannot: its key-makers are fixed.

\textbf{A hypothesis for the agreement.} The total tracks GPQA at 0.98 (\S{}4.5). GPQA is an accepted measure of capability with no theory of intelligence behind it: expert-written questions and a key (Rein et al., 2023). Of the eight constructs the Metanym Game demands, four are analogy under cognitive science's own names (Appendix E). Two instruments whose methods share nothing on the surface agree at 0.98, so the common thing is in the depth. Our hypothesis is that a language model holds its knowledge as archetypal contexts, each instantiated in the topic domains where it applies; the metanym game reproduces this organisation, and GPQA runs on it (Figure \ref{fig-mechanism}). The argument: training a language model is compression (Shannon, 1951; Del\'{e}tang et al., 2024), and analogy is semantic compression, a shared relational structure with different particulars (Gentner, 1983). The archetypal context is that structure in the latent space, the context template its literal representation, and the metanym sets the particulars that instantiate it. In the Metanym Game an instantiation is between two and a half and eleven times the size of its metanym set, so each domain is added for a fraction of the text it yields. This metanymic compression, in literal space, is of the order measured for language models, about ten (Del\'{e}tang et al., 2024), and above gzip's three. A memorised string compresses only its exact repeats; a frame compresses every new instance that fits it. The archetypal compression, in latent space, is greater still, since one archetype stands behind many context templates: between runs the models rewrite their templates and keep their archetypes (\S{}4.6). A form this compressive qualifies as a learning mechanism. GPQA and the Metanym Game then run the same mechanism from opposite ends. GPQA gives the archetype and the domain and four candidates: in 99.5\% of the replies the model derives the solution and then picks the candidate that matches it, the mechanism in use. The game gives nothing but rating criteria, truth among them: the model selects an archetype and domains and writes them out, the mechanism itself. This suggests that models will play the Metanym Game well without reasoning because the answer is retrieved, not built, and a reasoning channel should add little.

\textbf{A test.} We played four models each as two players, 1) with reasoning off at temperature 0, and 2) with reasoning on: Claude Sonnet 4.6, Claude Haiku 4.5, GPT-5.6 Terra and GPT-5.6 Luna. The thinking model does not build templates in its thinking; it writes them once, in the answer, as the non-thinking model does. What the thinking contains is a list of archetypes, by name, and a choice among them: selection over things the model already has, which is what retrieval looks like. The ratings agree (Table \ref{tab-thinking}): less than a point of change for every model, none clearing its interval. But for GPQA it is different: here thinking is helpful. Anthropic reports Opus 4 and Sonnet 4 at 74.9 and 70.0 without extended thinking and 79.6 and 75.4 with it (Anthropic, 2025); ours, reasoning off, read 71.2 and 72.2. That is the split the hypothesis predicts: for GPQA a derivation, which reasoning helps (Sprague et al., 2025); for the Metanym Game a retrieval, where reasoning is of much less help.

\begin{table}[htb]
\caption{Factual rating with the reasoning channel off and on, six judges each, anchor at 7; bootstrap over judges and archetypes, percentile 95\% intervals. One portfolio per cell; the vendors return summaries of the thinking, not the trace.}
\label{tab-thinking}
\begin{center}\small
\begin{tabular}{@{}lrrl@{}}
\toprule\noalign{}
Model & Reasoning off & Reasoning on & On minus off, 95\% interval \\
\midrule\noalign{}
Claude Haiku 4.5 & 5.99 & 5.31 & \ensuremath{-}0.67 {[}\ensuremath{-}1.83, +0.42{]} \\
GPT-5.6 Luna & 7.63 & 7.81 & +0.18 {[}\ensuremath{-}0.08, +0.47{]} \\
Claude Sonnet 4.6 & 7.00 & 7.31 & +0.31 {[}\ensuremath{-}0.09, +0.76{]} \\
GPT-5.6 Terra & 7.01 & 7.55 & +0.54 {[}\ensuremath{-}0.17, +1.31{]} \\
\bottomrule
\end{tabular}
\end{center}
\end{table}

\textbf{Predictions.}

\begin{itemize}
\tightlist
\item
  \textbf{Domain-matched agreement.} If the hypothesis is correct, the agreement should hold within each topic domain. A model's factual score on its biology parallel contexts should track its GPQA accuracy on the biology questions, and the same for physics and chemistry. GPQA labels each question by domain and the released evaluations record the domain of each parallel context, so the test needs no new run. If agreement within a domain is no higher than agreement across domains, the hypothesis is refuted.
\item
  \textbf{Separability.} If the hypothesis is correct, archetype and topic domain are separate in the latent space. The same archetype should be recoverable from its parallel contexts in unrelated domains, and the same domain from the parallel contexts of unrelated archetypes. If representations separate by domain only, the hypothesis is refuted.
\end{itemize}

What the data establish is narrower than the hypothesis: the correlation itself, 0.94 from the two factual quarters alone and 0.98 for the total (Appendix D.1).

\section{Limitations}\label{limitations}

\textbf{Steering signal, and its caveat.} Self-improvement, the council governing its own rules, is specified but not exercised. A system optimised against \(T\) is optimised against a consensus it participates in, so gains can come from courting the consensus; the partial answers are the two quarters consensus does not own and independently constituted councils.

\textbf{Scope.} The runs share one configuration --- one prompt template, one roster --- so the bootstrap intervals measure item and run-to-run dispersion, not the configuration (the four anchor values give the same leaderboard, Spearman 0.90--0.96); the leading group sits at its discrimination floor; a quarter of a non-council model's total rests on the contest's easier consistency test; no seat has yet been contested. Peer consensus is conservative against anomaly: a synthetic evaluator that reproduces the consensus and then inverts a fifth of its verdicts loses most of its competence (Appendix A.7), a penalty a dissenter pays on only half of \(T\).

\section{Conclusion}\label{conclusion}

The \emph{metanym game} is a structural test of intelligence built entirely of analogy, falsifiable sentence by sentence. The \emph{council-of-peers benchmark} needs nothing outside itself: truth as the dominant axis of inter-evaluator agreement, reliability as invariance under a swept anchor, judges certified by the participants, contestable seats --- and one external check, by design, \(r = 0.98\) against GPQA Diamond.\label{endmain}

\subsection*{AI use statement}

Every original idea in this work is the author's. The work was developed in a sustained dialogue with generative AI assistants (Anthropic's Claude), directed by the author throughout. Within that dialogue the assistants criticised the theory, the estimators and the experimental design as a reviewer in the field would, and proposed hypotheses and experiments alongside the author's; none was adopted without a competing hypothesis and a further experiment to test it. They implemented the analysis methods, cleaned and reformatted the evaluation records and GPQA response logs, and helped interpret results; they were also used for code, literature search, figures, references and first drafts, which the author rewrote into the author's own text. No data and no proofs were generated by AI: every rating in the paper comes from the twelve rated models of \S{}3.1. All AI-assisted output was checked against the deterministic re-derivation of every published number (Reproducibility statement). The author takes responsibility for the final content.

\subsection*{Reproducibility statement}

Everything behind this paper is released, as it was produced, in one package (https://github.com/dnordfors/metanym-game-paper, \texttt{reproduce/}): the prompts as sent; every submission the twelve models generated in the three runs; every rating every judge gave, in the three runs and the anchor sweep, with its written justification and the record of the API call; and every raw GPQA reply. From these records one command, \texttt{reproduce.sh}, re-derives every number, table and figure deterministically, each step labelled with the exhibit it produces, and a second, \texttt{verify\_\allowbreak{}chain.py}, confirms the records form one chain: each run names the prompt it was sent, each judgement names the submission it scored, each API record fits its transcript, each GPQA reply re-scores to the table it enters. A checksum manifest fixes the package's contents; its SHA-256 is \texttt{516b2571ca1fb99ae6744b698da28199ce9bdb37b787cf02e7773e6707cfd3aa}. The re-analysis makes no API calls and needs no credentials; it takes about a minute. The estimators are specified to the equation in Appendix A, including every convention (self-entry filling, row-centering, sign and clamp, Procrustes alignment of bootstrap replicates, the joint (run, submission, archetype) resampling grid). Producing a \emph{new} run --- re-querying the models --- is deliberately not part of the package: it costs API budget and is non-deterministic by construction; \S{}4.6 reports what moves across three independent regenerations.

\section*{References}
\begingroup\small\setlength{\parindent}{-1.5em}\setlength{\leftskip}{1.5em}

Anthropic (2025). Introducing Claude 4. Announcement, 22 May 2025. https://www.anthropic.com/news/claude-4

Bai, Y., et al.~(2023). Benchmarking foundation models with Language-Model-as-an-Examiner. \emph{NeurIPS 36.} arXiv:2306.04181.

Bellibatlu, R. R., Raff, E., \& Zhang, W. (2026). JudgeSense: A benchmark for prompt sensitivity in LLM-as-a-judge systems. arXiv:2604.23478.

Bonacich, P. (1972). Factoring and weighting approaches to status scores and clique identification. \emph{Journal of Mathematical Sociology, 2}(1), 113--120.

Cattell, R. B. (1963). Theory of fluid and crystallized intelligence: A critical experiment. \emph{Journal of Educational Psychology, 54}(1), 1--22.

Chang, J., Piff, L., Sana, S., Li, J. X., \& Levine, L. (2026). EigenBench: A comparative behavioral measure of value alignment. \emph{ICLR 2026.} arXiv:2509.01938.

Chollet, F. (2019). On the measure of intelligence. arXiv:1911.01547.

Dawid, A. P., \& Skene, A. M. (1979). Maximum likelihood estimation of observer error-rates using the EM algorithm. \emph{Journal of the Royal Statistical Society: Series C, 28}(1), 20--28.

Del\'{e}tang, G., Ruoss, A., Duquenne, P.-A., Catt, E., Genewein, T., Mattern, C., Grau-Moya, J., Wenliang, L. K., Aitchison, M., Orseau, L., Hutter, M., \& Veness, J. (2024). Language modeling is compression. In \emph{International Conference on Learning Representations (ICLR 2024)}. arXiv:2309.10668.

Don-Yehiya, S., Yehudai, A., Choshen, L., \& Abend, O. (2026). Mediocrity is the key for LLM as a judge anchor selection. \emph{ACL 2026.} arXiv:2603.16848.

Efron, B., \& Tibshirani, R. J. (1993). \emph{An Introduction to the Bootstrap.} Chapman \& Hall.

Falkenhainer, B., Forbus, K. D., \& Gentner, D. (1989). The structure-mapping engine: Algorithm and examples. \emph{Artificial Intelligence, 41}(1), 1--63.

Gentner, D. (1983). Structure-mapping: A theoretical framework for analogy. \emph{Cognitive Science, 7}(2), 155--170.

Guilford, J. P. (1967). \emph{The Nature of Human Intelligence.} McGraw-Hill.

Hesse, M. (1963). \emph{Models and Analogies in Science.} Sheed \& Ward.

Hofstadter, D., \& Sander, E. (2013). \emph{Surfaces and Essences.} Basic Books.

Horn, J. L., \& Cattell, R. B. (1966). Refinement and test of the theory of fluid and crystallized general intelligences. \emph{Journal of Educational Psychology, 57}(5), 253--270.

Ili\'{c}, D., \& Gignac, G. E. (2024). Evidence of interrelated cognitive-like capabilities in large language models: Indications of artificial general intelligence or achievement? \emph{Intelligence, 106}, 101858.

Kamvar, S. D., Schlosser, M. T., \& Garcia-Molina, H. (2003). The EigenTrust algorithm for reputation management in P2P networks. \emph{WWW 2003}, 640--651.

Lewis, M., \& Mitchell, M. (2024). Using counterfactual tasks to evaluate the generality of analogical reasoning in large language models. \emph{CogSci 2024.} arXiv:2402.08955.

Li, X. L., Shrivastava, V., Li, S., Hashimoto, T., \& Liang, P. (2024). Benchmarking and improving generator-validator consistency of language models. \emph{ICLR 2024.} arXiv:2310.01846.

Longino, H. E. (1990). \emph{Science as Social Knowledge.} Princeton University Press.

Mitchell, M. (2021). Abstraction and analogy-making in artificial intelligence. \emph{Annals of the New York Academy of Sciences, 1505}(1), 79--101.

Neisser, U. (1979). The concept of intelligence. \emph{Intelligence, 3}(3), 217--227.

Ning, K.-P., Yang, S., Liu, Y.-Y., Yao, J.-Y., Liu, Z.-H., Tian, Y.-H., Song, Y., \& Yuan, L. (2025). PiCO: Peer review in LLMs based on consistency optimization. \emph{ICLR 2025.} arXiv:2402.01830.

Oh, J., Kim, E., Cha, I., \& Oh, A. (2024). The Generative AI Paradox in evaluation: What it can solve, it may not evaluate. \emph{EACL 2024 Student Research Workshop}, 248--257. arXiv:2402.06204.

Parisi, F., Strino, F., Nadler, B., \& Kluger, Y. (2014). Ranking and combining multiple predictors without labeled data. \emph{PNAS, 111}(4), 1253--1258.

Penn, D. C., Holyoak, K. J., \& Povinelli, D. J. (2008). Darwin's mistake: Explaining the discontinuity between human and nonhuman minds. \emph{Behavioral and Brain Sciences, 31}(2), 109--130.

Rein, D., Hou, B. L., Stickland, A. C., Petty, J., Pang, R. Y., Dirani, J., Michael, J., \& Bowman, S. R. (2023). GPQA: A graduate-level Google-proof Q\&A benchmark. arXiv:2311.12022.

Shannon, C. E. (1951). Prediction and entropy of printed English. \emph{Bell System Technical Journal, 30}(1), 50--64.

Sprague, Z., Yin, F., Rodriguez, J. D., Jiang, D., Wadhwa, M., Singhal, P., Zhao, X., Ye, X., Mahowald, K., \& Durrett, G. (2025). To CoT or not to CoT? Chain-of-thought helps mainly on math and symbolic reasoning. In \emph{International Conference on Learning Representations (ICLR 2025)}. arXiv:2409.12183.

Srivastava, A., et al.~(2023). Beyond the imitation game: Quantifying and extrapolating the capabilities of language models. \emph{TMLR.} arXiv:2206.04615.

Sternberg, R. J., Conway, B. E., Ketron, J. L., \& Bernstein, M. (1981). People's conceptions of intelligence. \emph{Journal of Personality and Social Psychology, 41}(1), 37--55.

Verga, P., et al.~(2024). Replacing judges with juries: Evaluating LLM generations with a panel of diverse models. arXiv:2404.18796.

von Bertalanffy, L. (1968). \emph{General System Theory.} George Braziller.

Webb, T., Holyoak, K. J., \& Lu, H. (2023). Emergent analogical reasoning in large language models. \emph{Nature Human Behaviour, 7}(9), 1526--1541.

Weng, S., Feng, Y., \& Xie, X. (2026). Beyond accuracy: Policy invariance as a reliability test for LLM safety judges. arXiv:2605.06161.

West, P., Lu, X., Dziri, N., Brahman, F., Li, L., Hwang, J. D., Jiang, L., Fisher, J., Ravichander, A., Chandu, K., Newman, B., Koh, P. W., Ettinger, A., \& Choi, Y. (2024). The Generative AI Paradox: ``What it can create, it may not understand.'' \emph{ICLR 2024.} arXiv:2311.00059.

Zhang, Q., Ning, M., Liu, Z., Wang, Y., Ye, J., Huang, Y., Yang, S., Chen, X., Song, Y., \& Yuan, L. (2025). UPME: An unsupervised peer review framework for multimodal large language model evaluation. \emph{CVPR 2025}, 9165--9174. arXiv:2503.14941.

Zheng, L., et al.~(2023). Judging LLM-as-a-Judge with MT-Bench and Chatbot Arena. \emph{NeurIPS 36.} arXiv:2306.05685.

\endgroup
\clearpage
\appendix

\section{Rating estimators}\label{rating-estimators}

Every rating comes from one object: the scores the participants produce when each model grades the others' portfolios, swept across the anchor. No external answer key is used.

\textbf{Panel, tasks, anchor.} Twelve models, indexed \(s,t\in\{1,\dots,12\}\), are each a \emph{submission} (its portfolio is graded) and an \emph{evaluator} (it grades the others). Every model was asked to grade every portfolio including its own, and those self-evaluations are released; but \textbf{no rating below uses a model's grade of its own portfolio} --- every estimator is \emph{leave-self-out}. A portfolio holds five \emph{archetypes}, each realised as five \emph{parallel contexts}. Scoring uses six axes on a 1--10 scale: a \emph{factual} axis (once per parallel context) and five \emph{non-factual} axes --- beauty, intelligence, instantiation-distinctness, impressive-length (once per archetype) and structural-diversity (once per portfolio). Every score is relative to the \textbf{anchor} --- the model \(a\) whose portfolio won the un-anchored initial selection --- declared to score 7 on every axis. The anchor value is swept, \(\theta\in\Theta=\{5,6,7,8\}\), and \(\theta^{*}=7\) is the production anchor. Write \(r_{t,s,x,u}(\theta)\) for evaluator \(t\)'s score of unit \(u\) of axis \(x\) of submission \(s\) at anchor \(\theta\). A model earns a generation rating \(G\) and an evaluation rating \(E\) (A1); \(E\) has two parts (A12).

\subsection{Generation rating}\label{generation-rating}

Evaluator \(t\)'s overall score of submission \(s\) averages within each axis, then across axes:

\[ o_{t,s}=\frac{1}{|\mathcal{X}|}\sum_{x\in\mathcal{X}}\frac{1}{|U_x|}\sum_{u\in U_x} r_{t,s,x,u}(\theta^{*}), \tag{A2}\]

and the six-axis generation rating is the leave-self-out mean over the council \(\mathcal{C}\),

\[ G_s=\frac{1}{|\mathcal{C}\setminus\{s\}|}\sum_{t\in\mathcal{C},\,t\ne s} o_{t,s}. \tag{A3}\]

Intervals: 95\% percentile bootstrap over the per-(submission, archetype) units; a gap \(G_s-G_{s'}\) is \emph{resolvable} when the paired bootstrap puts its interval clear of 0. The official leaderboard's \(G\) is the split form (A13), \(G=\tfrac12(G^{F}+G^{C})\), not (A3).

\subsection{Factual competence and generation factuality (SVD)}\label{factual-competence-and-generation-factuality-svd}

Stack the factual scores into \(F\) (evaluators \ensuremath{\times} instantiations), each entry the 1--10 rating used \textbf{directly}, no thresholding. Self-entries and the rare missing entries are set to the anchor value:

\[ F_{sj}=r_{s,j}\in\{1,\dots,10\}\qquad(\text{self-entries }=\theta^{*}=7). \tag{A5}\]

Centre each row (subtract the evaluator's mean, removing its leniency) and keep the leading triple:

\[ \tilde F = F-\bar r\,\mathbf 1^{\top}=U\Sigma V^{\top},\qquad \sigma_1,\quad u\equiv U_{\cdot 1},\quad v\equiv V_{\cdot 1}. \tag{A6}\]

The left singular vector is factual competence,

\[ f\equiv u,\qquad \text{signed so } \textstyle\sum_s f_s>0,\qquad f^{+}_s=\max(f_s,0), \tag{A7}\]

clamped at zero so an evaluator anti-correlated with the consensus carries no weight. Filling the self-entries with each evaluator's own row mean, so that they vanish under centering, leaves the \(E^{F}\) ranking unchanged, moves no loading by more than 0.04, and moves the total's agreement with GPQA by less than 0.01 (\texttt{scripts/self\_\allowbreak{}entry\_\allowbreak{}fill\_\allowbreak{}check.py}). Centering is essential: raw scores cluster at the anchor, so on the un-centred matrix the leading axis is the shared level and ranks the most lenient evaluators highest. Equivalently \(u\) is the leading eigenvector of the row-centred inter-evaluator Gram \(\tilde F\tilde F^{\top}\). Because every row of \(\tilde F\) sums to zero, \(v\) sums to zero and is oriented so that positive means factually stronger (\(\operatorname{corr}(v,\ \text{column means of }\tilde F)>0\)).

The \textbf{competence-weighted consensus rating} of instantiation \(j\) reads \(v\) back on the 1--10 scale,

\[ \hat r_j \;=\; C+\kappa\,v_j,\qquad C=\frac{\sum_s f^{+}_s\,\bar r_s}{\sum_s f^{+}_s},\quad \kappa=\sigma_1\,\frac{\sum_s f^{+}_s\,u_s}{\sum_s f^{+}_s}>0, \tag{A8}\]

the rank-one approximation of the competence-weighted mean rating (the two agree within 0.14 on the canonical run, \(r = 1.00\)). Averaging over a generator's own instantiations \(J_g\) gives the key-free \textbf{generation-factuality} rating

\[ G^{F}_{g} \;=\; \frac1{|J_g|}\sum_{j\in J_g}\hat r_j, \tag{A9}\]

already on the 1--10 scale: clean instantiations sit at \(v_j\approx0\), hence at \(C\approx7\). Its interval resamples the generator's own \(\hat r_j\) with the consensus \((C,\kappa,v)\) held fixed. The one assumption: the only thing the evaluators share is the truth --- a same-vendor bloc with a common bias would add a spurious shared component, which is why competence is read off a vendor-diverse panel with the shared-bias check of \S{}4.2.

\begin{table}[H]
\caption{Evaluator factual competence \(E^{F}\) (left singular vector; ``anchored'' \(= 7f/f_a\)) and generator factuality \(G^{F}\) (right vector, per generator), twelve-evaluator basis, three runs pooled (805 items). \ensuremath{\dagger} loading interval includes zero; no interval printed.}
\label{tab-criterion-a}
\begin{center}\small
\shrinktowidth{\begin{tabular}{@{}lrrlrl@{}}
\toprule\noalign{}
Model & \(E^{F}\) loading & \(E^{F}\) anchored & 95\% CI & \(G^{F}\) & 95\% CI \\
\midrule\noalign{}
gemini-3.1-pro &\cellcolor[HTML]{2195C0}\textcolor[HTML]{EEF3F8}{0.61}&\cellcolor[HTML]{24459C}\textcolor[HTML]{EEF3F8}{8.24}& {[}7.33, 9.30{]} &\cellcolor[HTML]{1F7BB6}\textcolor[HTML]{EEF3F8}{6.78}& {[}6.62, 6.89{]}  \\
claude-opus-4.5 &\cellcolor[HTML]{3BB0C3}\textcolor[HTML]{1A1A1A}{0.52}&\cellcolor[HTML]{2072B2}\textcolor[HTML]{EEF3F8}{7.00}& {[}7.00, 7.00{]} &\cellcolor[HTML]{2072B2}\textcolor[HTML]{EEF3F8}{7.00}& (anchor)  \\
claude-opus-4.0 &\cellcolor[HTML]{8DD3BA}\textcolor[HTML]{1A1A1A}{0.35}&\cellcolor[HTML]{50BCC2}\textcolor[HTML]{1A1A1A}{4.69}& {[}3.45, 6.04{]} &\cellcolor[HTML]{2075B3}\textcolor[HTML]{EEF3F8}{6.93}& {[}6.89, 6.97{]}  \\
claude-opus-4.1 &\cellcolor[HTML]{8DD3BA}\textcolor[HTML]{1A1A1A}{0.35}&\cellcolor[HTML]{52BCC1}\textcolor[HTML]{1A1A1A}{4.65}& {[}3.56, 5.96{]} &\cellcolor[HTML]{2073B2}\textcolor[HTML]{EEF3F8}{6.99}& {[}6.91, 7.07{]}  \\
gemini-2.5-flash &\cellcolor[HTML]{C1E7B5}\textcolor[HTML]{1A1A1A}{0.26}&\cellcolor[HTML]{8FD3B9}\textcolor[HTML]{1A1A1A}{3.47}& {[}1.62, 5.05{]} &\cellcolor[HTML]{1E87BB}\textcolor[HTML]{EEF3F8}{6.49}& {[}6.27, 6.68{]}  \\
claude-sonnet-4 &\cellcolor[HTML]{DCF1B2}\textcolor[HTML]{1A1A1A}{0.18}&\cellcolor[HTML]{CBEBB4}\textcolor[HTML]{1A1A1A}{2.37}& {[}1.53, 3.36{]} &\cellcolor[HTML]{2074B2}\textcolor[HTML]{EEF3F8}{6.96}& {[}6.92, 7.00{]}  \\
gpt-4.1-mini &\cellcolor[HTML]{F1F9B9}\textcolor[HTML]{1A1A1A}{0.10}&\cellcolor[HTML]{EAF7B1}\textcolor[HTML]{1A1A1A}{1.34\ensuremath{\dagger}}& --- &\cellcolor[HTML]{1D91C0}\textcolor[HTML]{EEF3F8}{6.26}& {[}6.07, 6.43{]}  \\
gpt-4.1-2025-04-14 &\cellcolor[HTML]{F6FCC6}\textcolor[HTML]{1A1A1A}{0.06}&\cellcolor[HTML]{F4FBC0}\textcolor[HTML]{1A1A1A}{0.79}& {[}0.39, 1.53{]} &\cellcolor[HTML]{1E84BA}\textcolor[HTML]{EEF3F8}{6.56}& {[}6.39, 6.69{]}  \\
gpt-4o &\cellcolor[HTML]{F9FDCC}\textcolor[HTML]{1A1A1A}{0.04}&\cellcolor[HTML]{F8FCCA}\textcolor[HTML]{1A1A1A}{0.47}& {[}0.28, 0.70{]} &\cellcolor[HTML]{37ACC3}\textcolor[HTML]{EEF3F8}{5.33}& {[}5.12, 5.53{]}  \\
gpt-4o-2024-08-06 &\cellcolor[HTML]{FBFDCF}\textcolor[HTML]{1A1A1A}{0.03}&\cellcolor[HTML]{FAFDCE}\textcolor[HTML]{1A1A1A}{0.35}& {[}0.18, 0.61{]} &\cellcolor[HTML]{36AAC3}\textcolor[HTML]{EEF3F8}{5.39}& {[}5.16, 5.62{]}  \\
gpt-4.1-nano &\cellcolor[HTML]{FCFED3}\textcolor[HTML]{1A1A1A}{0.02}&\cellcolor[HTML]{FBFDD0}\textcolor[HTML]{1A1A1A}{0.28\ensuremath{\dagger}}& --- &\cellcolor[HTML]{4AB9C3}\textcolor[HTML]{1A1A1A}{4.82}& {[}4.01, 5.38{]}  \\
gpt-4o-mini &\cellcolor[HTML]{FFFFD9}\textcolor[HTML]{1A1A1A}{-0.00}&\cellcolor[HTML]{FFFFD9}\textcolor[HTML]{1A1A1A}{-0.00\ensuremath{\dagger}}& --- &\cellcolor[HTML]{6CC6BE}\textcolor[HTML]{1A1A1A}{4.14}& {[}3.60, 4.69{]}  \\
\bottomrule
\end{tabular}}
\end{center}
\end{table}

\subsection{Rating consistency}\label{rating-consistency}

For evaluator \(s\) and axis \(x\), let \(v_{s,x}(\theta)\) be \(s\)'s axis-\(x\) scores across that axis's units at anchor \(\theta\), leave-self-out (50 units for the four per-archetype axes, 250 for factual, 10 for structural diversity; 55 / 275 / 11 for the anchor, whose own portfolio is not among the graded eleven). The per-axis consistency is the mean Pearson correlation over the anchor pairs on which both vectors are non-constant,

\[ r_{s,x}=\frac{1}{|\mathcal P_{s,x}|}\sum_{(\theta,\theta')\in\mathcal P_{s,x}}\operatorname{corr}\!\big(v_{s,x}(\theta),v_{s,x}(\theta')\big). \tag{A10}\]

The \textbf{collapsed} score that gates the council averages the four per-archetype non-factual axes into one value per (submission, archetype) before correlating,

\[ \bar r_s=\frac{1}{|\mathcal P_{s}|}\sum_{(\theta,\theta')\in\mathcal P_{s}}\operatorname{corr}\!\big(v_{s}(\theta),v_{s}(\theta')\big), \tag{A11}\]

so axis-idiosyncratic noise partially cancels (gemini-2.5-flash: row mean of Table \ref{tab-criterion-b} \ensuremath{\approx} 0.70, collapsed 0.81). Factual is A.2's job and is excluded; structural diversity, one score per portfolio, is too coarse and is excluded too. Design counts are met for eight of twelve evaluators; a missing grading call costs units, not correctness (dropped pairwise).

\begin{table}[H]
\caption{Anchor-sweep consistency per evaluator and axis (A10): mean pairwise Pearson correlation of scores across anchor values 5/6/7/8. It measures self-consistency, not accuracy; the factual column is distinct from the factual-competence loading. gpt-4o-mini's factual entry is undefined (near-zero variance). The initial council's collapsed values \(\bar r\) (A11) with 95\% CIs: gemini-3.1-pro 0.94 {[}0.92, 0.96{]}, claude-opus-4.5 0.93 {[}0.89, 0.96{]}, gemini-2.5-flash 0.81 {[}0.75, 0.86{]}, claude-opus-4.0 0.85 {[}0.80, 0.89{]}, claude-opus-4.1 0.87 {[}0.82, 0.91{]}; their factual loadings with CIs: 0.58 {[}0.56, 0.60{]}, 0.55 {[}0.54, 0.58{]}, 0.37 {[}0.31, 0.40{]}, 0.35 {[}0.32, 0.35{]}, 0.28 {[}0.27, 0.30{]}.}
\label{tab-criterion-b}
\begin{center}\small
\shrinktowidth{\begin{tabular}{@{}lrrrrrr@{}}
\toprule\noalign{}
Evaluator & factual & beauty & intelligence & distinctness & length & struct \\
\midrule\noalign{}
gemini-3.1-pro &\cellcolor[HTML]{243392}\textcolor[HTML]{EEF3F8}{0.88}&\cellcolor[HTML]{24439B}\textcolor[HTML]{EEF3F8}{0.83}&\cellcolor[HTML]{1F2F88}\textcolor[HTML]{EEF3F8}{0.90}&\cellcolor[HTML]{253695}\textcolor[HTML]{EEF3F8}{0.87}&\cellcolor[HTML]{1F2F88}\textcolor[HTML]{EEF3F8}{0.90}&\cellcolor[HTML]{22318D}\textcolor[HTML]{EEF3F8}{0.89} \\
claude-opus-4.5 &\cellcolor[HTML]{1F2F88}\textcolor[HTML]{EEF3F8}{0.90}&\cellcolor[HTML]{253996}\textcolor[HTML]{EEF3F8}{0.86}&\cellcolor[HTML]{243C98}\textcolor[HTML]{EEF3F8}{0.85}&\cellcolor[HTML]{24409A}\textcolor[HTML]{EEF3F8}{0.84}&\cellcolor[HTML]{1D2E83}\textcolor[HTML]{EEF3F8}{0.91}&\cellcolor[HTML]{243C98}\textcolor[HTML]{EEF3F8}{0.85} \\
claude-opus-4.1 &\cellcolor[HTML]{24439B}\textcolor[HTML]{EEF3F8}{0.83}&\cellcolor[HTML]{24409A}\textcolor[HTML]{EEF3F8}{0.84}&\cellcolor[HTML]{243C98}\textcolor[HTML]{EEF3F8}{0.85}&\cellcolor[HTML]{216AAE}\textcolor[HTML]{EEF3F8}{0.72}&\cellcolor[HTML]{24409A}\textcolor[HTML]{EEF3F8}{0.84}&\cellcolor[HTML]{1B2C7E}\textcolor[HTML]{EEF3F8}{0.92} \\
claude-opus-4.0 &\cellcolor[HTML]{2354A3}\textcolor[HTML]{EEF3F8}{0.78}&\cellcolor[HTML]{24439B}\textcolor[HTML]{EEF3F8}{0.83}&\cellcolor[HTML]{24409A}\textcolor[HTML]{EEF3F8}{0.84}&\cellcolor[HTML]{2076B4}\textcolor[HTML]{EEF3F8}{0.69}&\cellcolor[HTML]{24439B}\textcolor[HTML]{EEF3F8}{0.83}&\cellcolor[HTML]{253996}\textcolor[HTML]{EEF3F8}{0.86} \\
gpt-4.1-mini &\cellcolor[HTML]{253695}\textcolor[HTML]{EEF3F8}{0.87}&\cellcolor[HTML]{2257A5}\textcolor[HTML]{EEF3F8}{0.77}&\cellcolor[HTML]{225EA8}\textcolor[HTML]{EEF3F8}{0.75}&\cellcolor[HTML]{225BA6}\textcolor[HTML]{EEF3F8}{0.76}&\cellcolor[HTML]{24409A}\textcolor[HTML]{EEF3F8}{0.84}&\cellcolor[HTML]{234A9E}\textcolor[HTML]{EEF3F8}{0.81} \\
claude-sonnet-4 &\cellcolor[HTML]{2072B2}\textcolor[HTML]{EEF3F8}{0.70}&\cellcolor[HTML]{225EA8}\textcolor[HTML]{EEF3F8}{0.75}&\cellcolor[HTML]{2257A5}\textcolor[HTML]{EEF3F8}{0.77}&\cellcolor[HTML]{225EA8}\textcolor[HTML]{EEF3F8}{0.75}&\cellcolor[HTML]{2354A3}\textcolor[HTML]{EEF3F8}{0.78}&\cellcolor[HTML]{234A9E}\textcolor[HTML]{EEF3F8}{0.81} \\
gpt-4.1-2025-04-14 &\cellcolor[HTML]{CAEAB4}\textcolor[HTML]{1A1A1A}{0.24}&\cellcolor[HTML]{1E8BBD}\textcolor[HTML]{EEF3F8}{0.64}&\cellcolor[HTML]{2166AC}\textcolor[HTML]{EEF3F8}{0.73}&\cellcolor[HTML]{2262AA}\textcolor[HTML]{EEF3F8}{0.74}&\cellcolor[HTML]{24469D}\textcolor[HTML]{EEF3F8}{0.82}&\cellcolor[HTML]{24409A}\textcolor[HTML]{EEF3F8}{0.84} \\
gemini-2.5-flash &\cellcolor[HTML]{A4DCB7}\textcolor[HTML]{1A1A1A}{0.31}&\cellcolor[HTML]{2257A5}\textcolor[HTML]{EEF3F8}{0.77}&\cellcolor[HTML]{1E83B9}\textcolor[HTML]{EEF3F8}{0.66}&\cellcolor[HTML]{41B6C4}\textcolor[HTML]{1A1A1A}{0.50}&\cellcolor[HTML]{24469D}\textcolor[HTML]{EEF3F8}{0.82}&\cellcolor[HTML]{243C98}\textcolor[HTML]{EEF3F8}{0.85} \\
gpt-4.1-nano &\cellcolor[HTML]{1E87BB}\textcolor[HTML]{EEF3F8}{0.65}&\cellcolor[HTML]{1D8FBF}\textcolor[HTML]{EEF3F8}{0.63}&\cellcolor[HTML]{1E8BBD}\textcolor[HTML]{EEF3F8}{0.64}&\cellcolor[HTML]{5FC1C0}\textcolor[HTML]{1A1A1A}{0.44}&\cellcolor[HTML]{82CEBB}\textcolor[HTML]{1A1A1A}{0.37}&\cellcolor[HTML]{88D0BA}\textcolor[HTML]{1A1A1A}{0.36} \\
gpt-4o-2024-08-06 &\cellcolor[HTML]{279BC1}\textcolor[HTML]{EEF3F8}{0.59}&\cellcolor[HTML]{2A9EC1}\textcolor[HTML]{EEF3F8}{0.58}&\cellcolor[HTML]{33A7C2}\textcolor[HTML]{EEF3F8}{0.55}&\cellcolor[HTML]{7DCCBB}\textcolor[HTML]{1A1A1A}{0.38}&\cellcolor[HTML]{38ADC3}\textcolor[HTML]{1A1A1A}{0.53}&\cellcolor[HTML]{2076B4}\textcolor[HTML]{EEF3F8}{0.69} \\
gpt-4o &\cellcolor[HTML]{EBF7B1}\textcolor[HTML]{1A1A1A}{0.13}&\cellcolor[HTML]{9FD9B8}\textcolor[HTML]{1A1A1A}{0.32}&\cellcolor[HTML]{CAEAB4}\textcolor[HTML]{1A1A1A}{0.24}&\cellcolor[HTML]{EEF8B3}\textcolor[HTML]{1A1A1A}{0.12}&\cellcolor[HTML]{73C8BD}\textcolor[HTML]{1A1A1A}{0.40}&\cellcolor[HTML]{93D5B9}\textcolor[HTML]{1A1A1A}{0.34} \\
gpt-4o-mini & n/a &\cellcolor[HTML]{C7E9B4}\textcolor[HTML]{1A1A1A}{0.25}&\cellcolor[HTML]{88D0BA}\textcolor[HTML]{1A1A1A}{0.36}&\cellcolor[HTML]{BBE5B5}\textcolor[HTML]{1A1A1A}{0.27}&\cellcolor[HTML]{F1F9B9}\textcolor[HTML]{1A1A1A}{0.10}&\cellcolor[HTML]{CDEBB4}\textcolor[HTML]{1A1A1A}{0.23} \\
\bottomrule
\end{tabular}}
\end{center}
\end{table}

\subsection{The council and the total}\label{the-council-and-the-total}

The sweep keeps the anchor interior to the 1--10 scale. At its ends, 5 and 8, the room on one side of the anchor shrinks, so part of the residual disagreement between anchor values is compression of the scale rather than reordering; the consistency values are conservative to that extent.

A model is \textbf{reliable} when its competence sits clear of the inert band --- its 95\% bootstrap interval on \(f\) separates it from the near-zero cluster --- and \(\bar r_s\ge0.78\). Eight of twelve clear the consistency gate; five clear both, and that conjunction seats the council. Each evaluator index is rescaled so the anchor scores 7,

\[ E^{F}_s=7\,\frac{f_s}{f_a},\qquad E^{C}_s=7\,\frac{\bar r_s}{\bar r_a}, \tag{A12}\]

so the singular vector's arbitrary scale cancels. Generation splits the same way: \(G^{F}\) (A9) and the criterion quality \(G^{C}\), the council's mean over the five non-factual axes, \textbf{reliability-weighted} --- each judge's vote on axis \(x\) weighted by its own \(r_{t,x}\):

\[ G^{C}_s=\frac{1}{|\mathcal{X}_{5}|}\sum_{x\in\mathcal{X}_{5}}\frac{\sum_{t\in\mathcal{C}\setminus\{s\}}\max(r_{t,x},0)\,\bar r_{t,s,x}}{\sum_{t\in\mathcal{C}\setminus\{s\}}\max(r_{t,x},0)}, \tag{A12b}\]

with \(\bar r_{t,s,x}\) judge \(t\)'s mean score of \(s\) on axis \(x\) at the production anchor. Then

\[ G_s=\tfrac12\big(G^{F}_s+G^{C}_s\big),\qquad E_s=\tfrac12\big(E^{F}_s+E^{C}_s\big), \tag{A13}\]

\[ T_s=\tfrac12\big(G_s+E_s\big)=\tfrac14\big(G^{F}_s+G^{C}_s+E^{F}_s+E^{C}_s\big). \tag{A14}\]

\textbf{Intervals.} Every rating carries a 95\% percentile bootstrap (\(10^3\)--\(10^4\) replicates). Units: \(G\) --- the per-(run, submission, archetype) units; \(f\) --- the 805 parallel contexts of the pooled matrix, each replicate's top-2 left subspace Procrustes-aligned to the full sample before the leading loading is read (the leading axis is near-degenerate between the Anthropic and Google blocs); \(\bar r\) --- the 55 (submission, archetype) atoms of run 1's anchor sweep. \(E\) and \(T\) are \textbf{not} combined analytically: they are bootstrapped jointly on the 165-atom (run, submission, archetype) grid, the run-1 atoms of each draw feeding \(\bar r\), leave-self-out applied after the resample, every component including \(f_a,\bar r_a\) recomputed on each replicate. Because a regenerated portfolio moves all of its atoms together, the interval contains run-to-run dispersion as well as item dispersion: pooling three runs did not narrow the intervals of run 1 (mean width ratio 1.06), and the two-run pool sits at 1.01 (Appendix F). The council is held at its selected membership rather than re-selected inside each replicate (the gate is itself defined by a bootstrap interval), so the interval is conditional on that selection.

\subsection{Per-criterion generator quality versus evaluator consistency}\label{per-criterion-generator-quality-versus-evaluator-consistency}

For each non-factual criterion \(x\), \(G_{s,x}\) is the reliability-weighted council estimator (A12b) on that axis and \(E_{s,x}=7\,r_{s,x}/r_{a,x}\); both anchored to 7. Their agreement is the \textbf{anchored cosine},

\[ \mathrm{cos}(G,E)_x=\frac{\sum_{s}(G_{s,x}-7)\,(E_{s,x}-7)}{\sqrt{\sum_{s}(G_{s,x}-7)^2}\;\sqrt{\sum_{s}(E_{s,x}-7)^2}}, \tag{A15}\]

centred on the anchor point rather than each vector's mean so values are comparable across axes. It is a diagnostic and feeds nothing in \(T\).

\begin{table}[H]
\caption{Per-criterion generator quality \(G\) versus evaluator consistency \(E\), both anchored to claude-opus-4.5 (\ensuremath{\star}) = 7; last row the anchored cosine with its joint-bootstrap 95\% CI.}
\label{tab-per-criterion}
\begin{center}\small
\shrinktowidth{\begin{tabular}{@{}lrrrrrrrrrr@{}}
\toprule\noalign{}
Model & beauty \(G\) & beauty \(E\) & intel \(G\) & intel \(E\) & dist \(G\) & dist \(E\) & len \(G\) & len \(E\) & struct \(G\) & struct \(E\) \\
\midrule\noalign{}
\ensuremath{\star} claude-opus-4.5 &\cellcolor[HTML]{2072B2}\textcolor[HTML]{EEF3F8}{7.0}&\cellcolor[HTML]{2072B2}\textcolor[HTML]{EEF3F8}{7.0}&\cellcolor[HTML]{2072B2}\textcolor[HTML]{EEF3F8}{7.0}&\cellcolor[HTML]{2072B2}\textcolor[HTML]{EEF3F8}{7.0}&\cellcolor[HTML]{2072B2}\textcolor[HTML]{EEF3F8}{7.0}&\cellcolor[HTML]{2072B2}\textcolor[HTML]{EEF3F8}{7.0}&\cellcolor[HTML]{2072B2}\textcolor[HTML]{EEF3F8}{7.0}&\cellcolor[HTML]{2072B2}\textcolor[HTML]{EEF3F8}{7.0}&\cellcolor[HTML]{2072B2}\textcolor[HTML]{EEF3F8}{7.0}&\cellcolor[HTML]{2072B2}\textcolor[HTML]{EEF3F8}{7.0} \\
claude-opus-4.1 &\cellcolor[HTML]{206EB0}\textcolor[HTML]{EEF3F8}{7.1}&\cellcolor[HTML]{1F7BB5}\textcolor[HTML]{EEF3F8}{6.8}&\cellcolor[HTML]{216AAE}\textcolor[HTML]{EEF3F8}{7.2}&\cellcolor[HTML]{2072B2}\textcolor[HTML]{EEF3F8}{7.0}&\cellcolor[HTML]{2262AA}\textcolor[HTML]{EEF3F8}{7.4}&\cellcolor[HTML]{2498C1}\textcolor[HTML]{EEF3F8}{6.0}&\cellcolor[HTML]{1F7FB7}\textcolor[HTML]{EEF3F8}{6.7}&\cellcolor[HTML]{1E87BB}\textcolor[HTML]{EEF3F8}{6.5}&\cellcolor[HTML]{2166AC}\textcolor[HTML]{EEF3F8}{7.3}&\cellcolor[HTML]{225BA6}\textcolor[HTML]{EEF3F8}{7.6} \\
claude-opus-4.0 &\cellcolor[HTML]{2076B4}\textcolor[HTML]{EEF3F8}{6.9}&\cellcolor[HTML]{1F7FB7}\textcolor[HTML]{EEF3F8}{6.7}&\cellcolor[HTML]{2076B4}\textcolor[HTML]{EEF3F8}{6.9}&\cellcolor[HTML]{2072B2}\textcolor[HTML]{EEF3F8}{7.0}&\cellcolor[HTML]{2072B2}\textcolor[HTML]{EEF3F8}{7.0}&\cellcolor[HTML]{2DA1C2}\textcolor[HTML]{EEF3F8}{5.7}&\cellcolor[HTML]{1F7BB5}\textcolor[HTML]{EEF3F8}{6.8}&\cellcolor[HTML]{1D8FBF}\textcolor[HTML]{EEF3F8}{6.3}&\cellcolor[HTML]{2166AC}\textcolor[HTML]{EEF3F8}{7.3}&\cellcolor[HTML]{206EB0}\textcolor[HTML]{EEF3F8}{7.1} \\
claude-sonnet-4 &\cellcolor[HTML]{1D8FBF}\textcolor[HTML]{EEF3F8}{6.3}&\cellcolor[HTML]{2195C0}\textcolor[HTML]{EEF3F8}{6.1}&\cellcolor[HTML]{1D8FBF}\textcolor[HTML]{EEF3F8}{6.3}&\cellcolor[HTML]{1E8BBD}\textcolor[HTML]{EEF3F8}{6.4}&\cellcolor[HTML]{1E87BB}\textcolor[HTML]{EEF3F8}{6.5}&\cellcolor[HTML]{1E92C0}\textcolor[HTML]{EEF3F8}{6.2}&\cellcolor[HTML]{2195C0}\textcolor[HTML]{EEF3F8}{6.1}&\cellcolor[HTML]{2498C1}\textcolor[HTML]{EEF3F8}{6.0}&\cellcolor[HTML]{1E8BBD}\textcolor[HTML]{EEF3F8}{6.4}&\cellcolor[HTML]{1F7FB7}\textcolor[HTML]{EEF3F8}{6.7} \\
gemini-3.1-pro &\cellcolor[HTML]{2A9EC1}\textcolor[HTML]{EEF3F8}{5.8}&\cellcolor[HTML]{1F7BB5}\textcolor[HTML]{EEF3F8}{6.8}&\cellcolor[HTML]{2DA1C2}\textcolor[HTML]{EEF3F8}{5.7}&\cellcolor[HTML]{2262AA}\textcolor[HTML]{EEF3F8}{7.4}&\cellcolor[HTML]{2498C1}\textcolor[HTML]{EEF3F8}{6.0}&\cellcolor[HTML]{2166AC}\textcolor[HTML]{EEF3F8}{7.3}&\cellcolor[HTML]{33A7C2}\textcolor[HTML]{EEF3F8}{5.5}&\cellcolor[HTML]{2076B4}\textcolor[HTML]{EEF3F8}{6.9}&\cellcolor[HTML]{2DA1C2}\textcolor[HTML]{EEF3F8}{5.7}&\cellcolor[HTML]{2262AA}\textcolor[HTML]{EEF3F8}{7.4} \\
gemini-2.5-flash &\cellcolor[HTML]{35AAC3}\textcolor[HTML]{EEF3F8}{5.4}&\cellcolor[HTML]{1E92C0}\textcolor[HTML]{EEF3F8}{6.2}&\cellcolor[HTML]{33A7C2}\textcolor[HTML]{EEF3F8}{5.5}&\cellcolor[HTML]{33A7C2}\textcolor[HTML]{EEF3F8}{5.5}&\cellcolor[HTML]{1E92C0}\textcolor[HTML]{EEF3F8}{6.2}&\cellcolor[HTML]{69C5BE}\textcolor[HTML]{1A1A1A}{4.2}&\cellcolor[HTML]{2DA1C2}\textcolor[HTML]{EEF3F8}{5.7}&\cellcolor[HTML]{1D8FBF}\textcolor[HTML]{EEF3F8}{6.3}&\cellcolor[HTML]{2A9EC1}\textcolor[HTML]{EEF3F8}{5.8}&\cellcolor[HTML]{2072B2}\textcolor[HTML]{EEF3F8}{7.0} \\
gpt-4.1-mini &\cellcolor[HTML]{4BBAC3}\textcolor[HTML]{1A1A1A}{4.8}&\cellcolor[HTML]{1D8FBF}\textcolor[HTML]{EEF3F8}{6.3}&\cellcolor[HTML]{41B6C4}\textcolor[HTML]{1A1A1A}{5.0}&\cellcolor[HTML]{1E92C0}\textcolor[HTML]{EEF3F8}{6.2}&\cellcolor[HTML]{279BC1}\textcolor[HTML]{EEF3F8}{5.9}&\cellcolor[HTML]{1D8FBF}\textcolor[HTML]{EEF3F8}{6.3}&\cellcolor[HTML]{69C5BE}\textcolor[HTML]{1A1A1A}{4.2}&\cellcolor[HTML]{1E87BB}\textcolor[HTML]{EEF3F8}{6.5}&\cellcolor[HTML]{35AAC3}\textcolor[HTML]{EEF3F8}{5.4}&\cellcolor[HTML]{1F7FB7}\textcolor[HTML]{EEF3F8}{6.7} \\
gpt-4.1-2025-04-14 &\cellcolor[HTML]{3EB3C4}\textcolor[HTML]{1A1A1A}{5.1}&\cellcolor[HTML]{3BB0C3}\textcolor[HTML]{1A1A1A}{5.2}&\cellcolor[HTML]{41B6C4}\textcolor[HTML]{1A1A1A}{5.0}&\cellcolor[HTML]{2498C1}\textcolor[HTML]{EEF3F8}{6.0}&\cellcolor[HTML]{2498C1}\textcolor[HTML]{EEF3F8}{6.0}&\cellcolor[HTML]{1E92C0}\textcolor[HTML]{EEF3F8}{6.2}&\cellcolor[HTML]{88D0BA}\textcolor[HTML]{1A1A1A}{3.6}&\cellcolor[HTML]{1D8FBF}\textcolor[HTML]{EEF3F8}{6.3}&\cellcolor[HTML]{30A4C2}\textcolor[HTML]{EEF3F8}{5.6}&\cellcolor[HTML]{2072B2}\textcolor[HTML]{EEF3F8}{7.0} \\
gpt-4.1-nano &\cellcolor[HTML]{8DD3BA}\textcolor[HTML]{1A1A1A}{3.5}&\cellcolor[HTML]{3EB3C4}\textcolor[HTML]{1A1A1A}{5.1}&\cellcolor[HTML]{82CEBB}\textcolor[HTML]{1A1A1A}{3.7}&\cellcolor[HTML]{38ADC3}\textcolor[HTML]{1A1A1A}{5.3}&\cellcolor[HTML]{69C5BE}\textcolor[HTML]{1A1A1A}{4.2}&\cellcolor[HTML]{82CEBB}\textcolor[HTML]{1A1A1A}{3.7}&\cellcolor[HTML]{8DD3BA}\textcolor[HTML]{1A1A1A}{3.5}&\cellcolor[HTML]{B6E2B6}\textcolor[HTML]{1A1A1A}{2.8}&\cellcolor[HTML]{9FD9B8}\textcolor[HTML]{1A1A1A}{3.2}&\cellcolor[HTML]{B0E0B6}\textcolor[HTML]{1A1A1A}{2.9} \\
gpt-4o-2024-08-06 &\cellcolor[HTML]{8DD3BA}\textcolor[HTML]{1A1A1A}{3.5}&\cellcolor[HTML]{50BCC2}\textcolor[HTML]{1A1A1A}{4.7}&\cellcolor[HTML]{99D7B8}\textcolor[HTML]{1A1A1A}{3.3}&\cellcolor[HTML]{5ABFC0}\textcolor[HTML]{1A1A1A}{4.5}&\cellcolor[HTML]{6EC7BE}\textcolor[HTML]{1A1A1A}{4.1}&\cellcolor[HTML]{A4DCB7}\textcolor[HTML]{1A1A1A}{3.1}&\cellcolor[HTML]{99D7B8}\textcolor[HTML]{1A1A1A}{3.3}&\cellcolor[HTML]{73C8BD}\textcolor[HTML]{1A1A1A}{4.0}&\cellcolor[HTML]{AADEB7}\textcolor[HTML]{1A1A1A}{3.0}&\cellcolor[HTML]{2DA1C2}\textcolor[HTML]{EEF3F8}{5.7} \\
gpt-4o &\cellcolor[HTML]{93D5B9}\textcolor[HTML]{1A1A1A}{3.4}&\cellcolor[HTML]{C1E7B5}\textcolor[HTML]{1A1A1A}{2.6}&\cellcolor[HTML]{93D5B9}\textcolor[HTML]{1A1A1A}{3.4}&\cellcolor[HTML]{D6EFB3}\textcolor[HTML]{1A1A1A}{2.0}&\cellcolor[HTML]{50BCC2}\textcolor[HTML]{1A1A1A}{4.7}&\cellcolor[HTML]{F1F9B9}\textcolor[HTML]{1A1A1A}{1.0}&\cellcolor[HTML]{C1E7B5}\textcolor[HTML]{1A1A1A}{2.6}&\cellcolor[HTML]{A4DCB7}\textcolor[HTML]{1A1A1A}{3.1}&\cellcolor[HTML]{9FD9B8}\textcolor[HTML]{1A1A1A}{3.2}&\cellcolor[HTML]{B6E2B6}\textcolor[HTML]{1A1A1A}{2.8} \\
gpt-4o-mini &\cellcolor[HTML]{8DD3BA}\textcolor[HTML]{1A1A1A}{3.5}&\cellcolor[HTML]{D3EEB3}\textcolor[HTML]{1A1A1A}{2.1}&\cellcolor[HTML]{8DD3BA}\textcolor[HTML]{1A1A1A}{3.5}&\cellcolor[HTML]{AADEB7}\textcolor[HTML]{1A1A1A}{3.0}&\cellcolor[HTML]{99D7B8}\textcolor[HTML]{1A1A1A}{3.3}&\cellcolor[HTML]{CDEBB4}\textcolor[HTML]{1A1A1A}{2.3}&\cellcolor[HTML]{7DCCBB}\textcolor[HTML]{1A1A1A}{3.8}&\cellcolor[HTML]{F3FBBF}\textcolor[HTML]{1A1A1A}{0.8}&\cellcolor[HTML]{AADEB7}\textcolor[HTML]{1A1A1A}{3.0}&\cellcolor[HTML]{D9F0B3}\textcolor[HTML]{1A1A1A}{1.9} \\
cos(G,E) &\cellcolor[HTML]{F2FABC}\textcolor[HTML]{1A1A1A}{0.91}& {[}.83,.95{]} &\cellcolor[HTML]{F2FABC}\textcolor[HTML]{1A1A1A}{0.90}& {[}.80,.93{]} &\cellcolor[HTML]{F2FABD}\textcolor[HTML]{1A1A1A}{0.89}& {[}.82,.92{]} &\cellcolor[HTML]{F3FABE}\textcolor[HTML]{1A1A1A}{0.84}& {[}.80,.87{]} &\cellcolor[HTML]{F2FABD}\textcolor[HTML]{1A1A1A}{0.87}& {[}.86,.87{]}  \\
\bottomrule
\end{tabular}}
\end{center}
\end{table}

\subsection{The contest, the ballast, and the guards}\label{the-contest-the-ballast-and-the-guards}

The official leaderboard is issued on the \textbf{contest} basis: a contest for model \(c\) convenes the five council seats and \(c\) itself as evaluators over the incumbents' portfolios, the two ballast blocks and \(c\)'s own --- every estimator above unchanged, leave-self-out throughout, the anchor's \(f_a,\bar r_a\) the contest's own. Two \textbf{guards} decide whether the contest's factual axis is identified: \(\sigma_1/\sigma_2\in[2.0,5.0]\) and a spread of the seats' anchored \(E^{F}\) above 2.5 points; if either fails, no seat changes hands and the contest's totals carry the caveat. The upper bound marks the range within which the sizing below was validated, not a failure in itself.

A contest convenes the top of the field, and with the weak submissions gone the factual axis breaks: the seats' anchored \(E^{F}\) come out wrong in scale and in order, swinging by up to 8.8 points with the contestant. The \textbf{ballast} --- the weakest archived submissions, added to every contest's graded set and held fixed through anchor replacements and council rotations --- repairs this; two suffice.

\begin{table}[H]
\caption{Each seat's anchored \(E^{F}\) by contest composition --- 0--3 ballast blocks (mean over the seven possible contestants) beside the twelve-participant reference, three runs pooled. Council alone, the column is scrambled; from two ballast on, the contest reproduces the reference (mean \(|\Delta|\) 0.60 over the seven contests at two blocks, 0.37 at three; the same seat lowest in six of the seven). Two blocks is the protocol's configuration.}
\label{tab-ballast}
\begin{center}\small
\shrinktowidth{\begin{tabular}{@{}lrrrrr@{}}
\toprule\noalign{}
Seat & council alone & +1 ballast & +2 ballast & +3 ballast & all 12 \\
\midrule\noalign{}
gemini-3.1-pro &\cellcolor[HTML]{216BAE}\textcolor[HTML]{EEF3F8}{7.19}&\cellcolor[HTML]{2075B3}\textcolor[HTML]{EEF3F8}{6.94}&\cellcolor[HTML]{2353A3}\textcolor[HTML]{EEF3F8}{7.84}&\cellcolor[HTML]{24499E}\textcolor[HTML]{EEF3F8}{8.13}&\cellcolor[HTML]{24459C}\textcolor[HTML]{EEF3F8}{8.24} \\
\ensuremath{\star} claude-opus-4.5 &\cellcolor[HTML]{2072B2}\textcolor[HTML]{EEF3F8}{7.00}&\cellcolor[HTML]{2072B2}\textcolor[HTML]{EEF3F8}{7.00}&\cellcolor[HTML]{2072B2}\textcolor[HTML]{EEF3F8}{7.00}&\cellcolor[HTML]{2072B2}\textcolor[HTML]{EEF3F8}{7.00}&\cellcolor[HTML]{2072B2}\textcolor[HTML]{EEF3F8}{7.00} \\
claude-opus-4.0 &\cellcolor[HTML]{2073B2}\textcolor[HTML]{EEF3F8}{6.98}&\cellcolor[HTML]{89D1BA}\textcolor[HTML]{1A1A1A}{3.58}&\cellcolor[HTML]{84CFBA}\textcolor[HTML]{1A1A1A}{3.66}&\cellcolor[HTML]{6CC6BE}\textcolor[HTML]{1A1A1A}{4.14}&\cellcolor[HTML]{50BCC2}\textcolor[HTML]{1A1A1A}{4.69} \\
claude-opus-4.1 &\cellcolor[HTML]{2261AA}\textcolor[HTML]{EEF3F8}{7.42}&\cellcolor[HTML]{66C4BF}\textcolor[HTML]{1A1A1A}{4.26}&\cellcolor[HTML]{7DCCBB}\textcolor[HTML]{1A1A1A}{3.79}&\cellcolor[HTML]{69C5BE}\textcolor[HTML]{1A1A1A}{4.20}&\cellcolor[HTML]{52BCC1}\textcolor[HTML]{1A1A1A}{4.65} \\
gemini-2.5-flash &\cellcolor[HTML]{E8F6B1}\textcolor[HTML]{1A1A1A}{1.41}&\cellcolor[HTML]{B9E3B5}\textcolor[HTML]{1A1A1A}{2.75}&\cellcolor[HTML]{6AC5BE}\textcolor[HTML]{1A1A1A}{4.18}&\cellcolor[HTML]{75C9BC}\textcolor[HTML]{1A1A1A}{3.95}&\cellcolor[HTML]{8FD3B9}\textcolor[HTML]{1A1A1A}{3.47} \\
\bottomrule
\end{tabular}}
\end{center}
\end{table}

Why two and not one: with a single block the axis narrows toward \emph{did this judge notice the one bad portfolio}, and the guards fail in 6\% of bootstrap resamples; two blocks carry two independent error patterns, the guards hold in every resample, \(\sigma_1/\sigma_2 = 2.87\) {[}2.50, 3.03{]} on the pooled corpus with both interval ends inside the band; a third tightens the seats' fidelity further (0.37) at twenty-five more columns of grading per run, and the protocol keeps two. Taken apart, the runs behave differently: the sizing holds on run 1 (two-ballast separation 2.99 {[}2.71, 3.48{]}, guards holding in every resample), is marginal on run 2 (2.01 {[}1.87, 2.40{]}, holding in 65\% of resamples), and fails on run 3 at every ballast size (1.53 {[}1.31, 1.73{]}, holding in none): no column set can single out an axis the judges did not share. Against a per-evaluator permutation null (each evaluator's ratings shuffled across the columns), \(\sigma_1\) stands 1.48\ensuremath{\times} above the 95th-percentile noise edge on the pooled corpus, where the second pattern lies within noise; on run 3 alone it stands 1.29\ensuremath{\times} above the edge and the second pattern also exceeds it, while the leading direction is stable under the column bootstrap (\texttt{scripts/spectral\_\allowbreak{}gap\_\allowbreak{}checks.py}): run 3 has a shared factual axis, but not a single one, and pooling it with two clean runs restores one.

\subsection{Authority versus consistency on the subjective axes}\label{authority-versus-consistency-on-the-subjective-axes}

Computed on run 1 alone. Peer centrality can be run on the subjective axes too: one SVD per non-factual axis yields an \textbf{authority} rating --- each judge's alignment with the participants' collective taste. On the five subjective axes authority and consistency correlate at Pearson 0.70--0.90; on factual --- the one axis with a truth to be right about --- they diverge (0.50, CI {[}\ensuremath{-}0.08, 0.84{]}), because only there can a judge be stable yet wrong. Consistency credits a judge's whole stable standard, personal taste included, penalising only flimsiness; authority credits the collective share alone, so stable-but-partly-private judges drop under authority (Spearman between the twelve orderings 0.83). Council membership is invariant to the choice: the top five by either estimator are the five seats. Substituting authority for consistency in the total preserves the ordering (Spearman 0.986) with one headline change: authority is not bounded by the anchor's own loading, so gemini-3.1-pro's total (7.86) overtakes the pinned 7. The official rating uses consistency; adopting authority inside \(T\) would convert alignment into authority on axes where no truth licenses the conversion (\S{}6).

\textbf{Limits of a consensus-defined competence} (run 1). Because \(E^{F}\) is read off agreement, a judge that departs from the panel is scored down whether it is wrong or right. A synthetic evaluator built to reproduce the panel's competence-weighted consensus exactly reads \(E^{F} = 5.15\); inverting its verdict on 5\% of the items drops it to 4.84, on 20\% to 3.79 (\texttt{scripts/consensus\_\allowbreak{}limits.py}). Substituting authority for consistency in the total preserves the ordering (Spearman 0.986) save Gemini 3.1 Pro overtaking the anchor (7.86).

\subsection{The two estimators compared}\label{the-two-estimators-compared}

\begin{table}[H]
\caption{The two estimators and their division of labour (source \S{}4.2).}
\label{tab-estimators}
\begin{center}\footnotesize
\begin{tabular}{>{\raggedright\arraybackslash}p{0.14\linewidth}>{\raggedright\arraybackslash}p{0.37\linewidth}>{\raggedright\arraybackslash}p{0.41\linewidth}}
\toprule\noalign{}
& Peer centrality (graded SVD) & Rating consistency (anchor sweep) \\
\midrule\noalign{}
Character & \emph{collective}: each judge weighted by the other judges' agreement with it & \emph{individual}: each judge measured only against itself across the sweep \\
Licensing assumption & the only thing competent judges share is the truth & a stable standard is the only competence a subjective axis can show \\
Use for & fact-checking --- replaces the golden key (\(E^{F}\), \(G^{F}\)) & axes with no right or wrong (\(E^{C}\); the per-axis weights in \(G^{C}\)) \\
Blind spot & a misunderstanding \emph{shared} by the judges reads as truth (\S{}6) & a \emph{private} misconception, held consistently, passes as a standard \\
Why it stops there & agreement on taste would convert alignment into authority (A.7) & consistency cannot certify truth: a consistent judge can be consistently \emph{wrong} \\
\bottomrule
\end{tabular}
\end{center}
\end{table}

\section{Generation and evaluation prompts}\label{generation-and-evaluation-prompts}

The verbatim prompts used in the canonical run of \S{}4 of the main text. All are run with Temperature = 0, reasoning disabled, and tools disabled.

\begin{itemize}
\tightlist
\item
  \textbf{B.1} is the generation prompt: each model produces its five-archetype portfolio from it.
\item
  \textbf{B.2} is the evaluation prompt, shown in its \textbf{calibrated/anchored} form --- the version used for the anchored re-evaluation and the official council ratings (\S{}4.2--\S{}4.4) and in the steady-state protocol (\S{}4.3). It scores one \emph{Target} submission against a fixed \emph{Reference} submission pinned at \texttt{\{ANCHOR\_\allowbreak{}SCORE\}} on every criterion.
\end{itemize}

The bootstrap's initial all-against-all selection (\S{}4.1) uses the \textbf{un-anchored} form of the same prompt --- identical six criteria and JSON schema, with the calibration machinery removed. The exact passages that are absent in the un-anchored bootstrap pass are listed in the \textbf{Bootstrap note} after B.2, so both forms are fully specified from the single prompt below.

Template variables appear in braces: \texttt{\{SUBMISSIONS\}}, \texttt{\{REFERENCE\_\allowbreak{}SUBMISSION\}}, \texttt{\{TARGET\_\allowbreak{}SUBMISSION\}}, and \texttt{\{ANCHOR\_\allowbreak{}SCORE\}} (swept across \{5, 6, 7, 8\}; fixed at 7 for the official ratings).

\begin{center}\rule{0.5\linewidth}{0.5pt}\end{center}

\subsection{--- Generation prompt}\label{generation-prompt}

\begin{lstlisting}
# Make more of these. This is a contest -- your submissions will be ranked.

You will propose new **archetypal contexts** -- universal relational templates
that apply across multiple distant domains. Below are two worked examples,
then your task.

---

## Terminology

- **Archetypal context**: an essential context in its purest abstraction.
- **Context template**: a worded template with `[SLOT]` representing an archetypal context.
- **Parallel contexts** (also called *metaphors*): contexts that are instantiations of the same archetypal context / context template.
- **Metanyms**: words that mirror each other across parallel contexts without being synonyms.
- **Metanym set**: the set of metanyms that instantiates the context-template, producing one parallel context.
- **Metanym table**: the table whose columns are the metanym sets of the parallel contexts.

---

## Example 1

### Template

"[SIGNALING] is part of a complex system of communication that governs basic [ELEMENT] activities and coordinates [ELEMENT] actions. The ability of [ELEMENT] to perceive and correctly respond to [BOUNDARY] is the basis of development, [SUBSYSTEM] repair, and [RESILIENCE] as well as normal [SUBSYSTEM] [HOMEOSTASIS]. Errors in [ELEMENT] information processing are responsible for [FAILURE]. By understanding [SIGNALING], [FAILURE] may be treated effectively. [KNOWLEDGE SYSTEM] research helps us to understand the underlying structure of [SIGNALING] networks. [SIGNALING] is mostly thought of as signaling between [ELEMENT] of a single [SYSTEM]. However, [SIGNALING] may also occur between the [ELEMENT] of two different [SYSTEM]."

### Substitution table (metanyms in base form)

| [SLOT]            | Cell Signaling   | Organ Signaling          | Human Language   |
|-------------------|------------------|--------------------------|------------------|
| ELEMENT           | cell             | organ                    | human            |
| SIGNALING         | cell signaling   | endocrine signaling      | human language   |
| SUBSYSTEM         | tissue           | organ system             | community        |
| RESILIENCE        | immunity         | physiological resilience | resilience       |
| HOMEOSTASIS       | homeostasis      | systemic homeostasis     | equilibrium      |
| BOUNDARY          | microenvironment | internal environment     | environment      |
| FAILURE           | disease          | organ failure            | dysfunction      |
| KNOWLEDGE SYSTEM  | systems biology  | physiology               | sociology        |
| SYSTEM            | organism         | organism                 | society          |

### Cell Signaling

**Form (a)** -- grammatical substitution (metanyms inflected as English requires):
"Cell signaling is part of a complex system of communication that governs basic cell activities and coordinates cell actions. The ability of cells to perceive and correctly respond to their microenvironment is the basis of development, tissue repair, and immunity, as well as normal tissue homeostasis. Errors in cellular information processing are responsible for disease. By understanding cell signaling, disease may be treated effectively. Systems biology research helps us to understand the underlying structure of cell-signaling networks. Cell signaling is mostly thought of as signaling between cells of a single organism. However, cell signaling may also occur between the cells of two different organisms."

**Form (b)** -- idiomatic rewrite (same propositions, written as a domain expert would):
"Cell signaling is the communication apparatus that governs and coordinates cellular behavior. A cell's ability to sense and respond appropriately to its microenvironment underlies development, tissue repair, immunity, and ordinary tissue homeostasis. When that information processing fails, disease results -- and conversely, a clear understanding of cell signaling enables effective therapeutic intervention. Systems biology unpacks the structure of these signaling networks. Most cell signaling occurs within a single organism, but inter-organism signaling (host-pathogen, microbiome) is well-documented."

(Two more domains would follow with their own form (a) and form (b).)

---

## Example 2

### Template

"A [AGENT] must commit [RESOURCE] under uncertainty, and once a [COMMITMENT] is observed it cannot be costlessly reversed. As [INFORMATION] arrives, the [AGENT] learns that earlier [COMMITMENT] are increasingly suboptimal. [REVERSAL_COST] grows with the depth of prior [COMMITMENT], so the [AGENT] often continues along the original [PATH] even when fresh [INFORMATION] favors a different one. [DECISION_THEORY] studies how rational [AGENT] balance the value of [INFORMATION] against the cost of [REVERSAL_COST]."

### Substitution table

| [SLOT]            | Capital Investment   | Coalition Politics   |
|-------------------|----------------------|----------------------|
| AGENT             | firm                 | coalition            |
| RESOURCE          | capital              | endorsement          |
| COMMITMENT        | investment           | public statement     |
| INFORMATION       | market signal        | polling data         |
| REVERSAL_COST     | switching cost       | reputational cost    |
| PATH              | strategy             | position             |
| DECISION_THEORY   | investment theory    | political science    |

### Capital Investment

**Form (a)**:
"A firm must commit capital under uncertainty, and once an investment has been made it cannot be costlessly reversed. As market signals arrive, the firm learns that earlier investments are increasingly suboptimal. Switching costs grow with the depth of prior investments, so the firm often continues along the original strategy even when fresh market signals favor a different one. Investment theory studies how rational firms balance the value of market signals against the cost of switching."

**Form (b)**:
"Capital investments must be made under uncertainty, and once committed they are sunk -- reversal is costly. New market signals continuously update what would have been optimal, but the depth of prior commitment raises the cost of changing course. Firms therefore tend to stay with their original strategy, even when current information would favor switching. Real-options theory and other strands of investment theory characterise how rational firms trade off information value against reversal cost."

---

## Your task

Propose **five archetypal contexts**. Each archetypal context has a worded context-template, one metanym table with five metanym sets, and five parallel contexts (the instantiations of the template). The five archetypal contexts in your submission should themselves have very different system structures from each other. Surface relabelings of the worked examples above don't count.

### Note

Example 1 is **recursive**: cells - organs - humans. Recursive archetypal contexts can be observed in nature. But not all archetypal contexts are recursive. You are free to submit archetypal contexts of both kinds. If there are recursive ones in your submission, point to them. The instantiations should demonstrate the recursion.

### What to submit

For each of your five archetypal contexts, begin with:

```
## Archetype Proposal: <short name>
```

Then provide, for that archetypal context:

1. **Context-template** -- a worded paragraph with `[SLOT]` placeholders. Slots use one canonical noun (e.g. `[ELEMENT]`, never `[ELEMENTS]`).
2. **Metanym table** -- rows = slots, columns = 5 domains, each cell a metanym in **base form** (singular noun, infinitive verb, etc.).
3. **Five parallel contexts**, one per domain:
   - **Form (a)** -- the context-template with that domain's metanym set substituted in. Inflect metanyms as English requires; Form (a) must be grammatically correct.
   - **Form (b)** -- idiomatic rewrite of Form (a). Same propositions, written as a domain expert would naturally write them.
   - **Optional (*@\ensuremath{\le}@*)1-sentence justification** beginning `Justification:` -- only if a propositional claim might be misread by a domain expert.

### Rules

- The **context-template** uses base-form slot placeholders -- `[ELEMENT]` not `[ELEMENTS]`. One token per slot, used consistently.
- The **metanym table** lists metanyms in **base form** -- `cell`, `human`, etc.
- The **parallel contexts** (Form (a) and Form (b)) must use the **correct grammatical form** of each metanym for the sentence -- `cell` in the table becomes `cells` or `cell's` in the PC as English grammar requires.
- Every proposition in Form (a) must appear in Form (b), and vice versa. Do not add or drop claims between the two forms.

### How you will be ranked

A submission contains **five archetypal contexts**. Evaluators score on six criteria, each rated 1-10. The scope tag tells you the unit of judgment:

1. **(Each parallel context)** Each sentence is factually correct
2. **(Each archetypal context)** Beauty
3. **(Each archetypal context)** Intelligence
4. **(Each archetypal context)** The parallel contexts from the template span very different domains. Metanyms are far from synonymous
5. **(Each archetypal context)** The archetypal template has impressive length
6. **(Each submitted set of archetypal contexts)** The archetypal contexts have very different system structures
\end{lstlisting}

\begin{center}\rule{0.5\linewidth}{0.5pt}\end{center}

\subsection{--- Evaluation prompt (calibrated/anchored)}\label{evaluation-prompt-calibratedanchored}

\begin{lstlisting}
# Score this submission against a calibration reference.

You are evaluating one contest submission ("Target Submission") against a fixed
reference ("Reference Submission") that has been pre-scored at **{ANCHOR_SCORE}/10 on every
criterion**. Score the Target Submission only -- the Reference is your yardstick.

For each criterion below, ask: *is the Target's quality on this criterion better
or worse than the Reference, and by how much?*

- Equal quality to the Reference (*@\ensuremath{\rightarrow}@*) **{ANCHOR_SCORE}**
- Clearly better than the Reference (*@\ensuremath{\rightarrow}@*) **above {ANCHOR_SCORE}** (with magnitude reflecting how much better, up to 10)
- Clearly worse than the Reference (*@\ensuremath{\rightarrow}@*) **below {ANCHOR_SCORE}** (with magnitude reflecting how much worse, down to 1)

Use the full 1-10 scale relative to the calibration anchor. Do not score the
Reference Submission itself -- its scores are fixed at {ANCHOR_SCORE}.

## Terminology

- **Archetypal context**: an essential context in its purest abstraction.
- **Context template**: a worded template with `[SLOT]` representing an archetypal context.
- **Parallel contexts** (also called *metaphors*): contexts that are instantiations of the same archetypal context / context template.
- **Metanyms**: words that mirror each other across parallel contexts without being synonyms.
- **Metanym set**: the set of metanyms that instantiates the context-template, producing one parallel context.
- **Metanym table**: the table whose columns are the metanym sets of the parallel contexts.

---

Each submission contains **five archetypal contexts**. Each archetypal context has:

- A **context-template** -- a worded paragraph with `[SLOT]` placeholders.
- A **metanym table** -- five metanym sets, one per parallel context. Rows = slots, columns = domains.
- **Five parallel contexts** (the five instantiations of the template), each consisting of:
  - **Form (a)** -- the template with one metanym set substituted in, grammatically correct.
  - **Form (b)** -- an idiomatic rewrite of Form (a), same propositions in domain-expert prose.
  - Optionally a **Justification** sentence.

Score the Target Submission on **six criteria**, each rated 1-10 relative to the Reference (which is fixed at {ANCHOR_SCORE} on every criterion). The scope tag at the start of each criterion -- `(Each parallel context)`, `(Each archetypal context)`, or `(Each submitted set of archetypal contexts)` -- tells you the unit of judgment. For each scored unit, write one paragraph justifying the rating relative to the Reference, then give the number.

---

## The six criteria

### 1. (Each parallel context) Each sentence is factually correct (1-10)

### 2. (Each archetypal context) Beauty (1-10)

### 3. (Each archetypal context) Intelligence (1-10)

### 4. (Each archetypal context) The parallel contexts from the template span very different domains. Metanyms are far from synonymous (1-10)

### 5. (Each archetypal context) The archetypal template has impressive length (1-10)

### 6. (Each submitted set of archetypal contexts) The archetypal contexts have very different system structures (1-10)

---

## Note on recursion

Some submissions may be **recursive** -- the same archetypal context manifesting at multiple nested scales (cells (*@\ensuremath{\rightarrow}@*) organs (*@\ensuremath{\rightarrow}@*) humans, the canonical example). Contestants are invited to identify recursion in their submission and show the instantiations that demonstrate it. Recursion is a valued property when present and correctly identified, but is not required. Take it into account where appropriate.

---

## The submissions

### Reference Submission (fixed at {ANCHOR_SCORE}/10 on every criterion)

{REFERENCE_SUBMISSION}

---

### Target Submission (to be scored relative to the Reference)

{TARGET_SUBMISSION}

---

## Output

Produce a section in this exact form (for the Target only -- do not re-score the Reference):

```
## Target Submission

### Archetypal context 1: <short name>

#### Factually correct (per parallel context)
- PC 1 (<domain>): <one paragraph, relative to Reference>. Rating: N
- PC 2 (<domain>): <one paragraph, relative to Reference>. Rating: N
- PC 3 (<domain>): <one paragraph, relative to Reference>. Rating: N
- PC 4 (<domain>): <one paragraph, relative to Reference>. Rating: N
- PC 5 (<domain>): <one paragraph, relative to Reference>. Rating: N

#### Beauty
<one paragraph relative to Reference>
Rating: N

#### Intelligence
<one paragraph relative to Reference>
Rating: N

#### Domains far apart / metanyms not synonymous
<one paragraph relative to Reference>
Rating: N

#### Impressive length
<one paragraph relative to Reference>
Rating: N

### Archetypal context 2: <short name>
... (same five blocks)

### Archetypal context 3: <short name>
...

### Archetypal context 4: <short name>
...

### Archetypal context 5: <short name>
...

### Structural diversity across the submitted set
<one paragraph relative to Reference>
Rating: N
```

After the markdown, end with a single fenced JSON block (Target scores only):

```json
{
  "scores": {
    "Target": {
      "archetypal_contexts": [
        {
          "name": "<short name>",
          "factual_per_pc":           [N, N, N, N, N],
          "beauty":                   N,
          "intelligence":             N,
          "instantiation_distinctness": N,
          "impressive_length":        N
        }
        /* five entries in this list, one per archetypal context */
      ],
      "structural_diversity": N
    }
  }
}
```

All ratings are integers 1-10 inclusive. Equal to the Reference = {ANCHOR_SCORE}.
\end{lstlisting}

\subsection{Bootstrap note --- the un-anchored form (\S{}4.1 initial selection)}\label{bootstrap-note-the-un-anchored-form-4.1-initial-selection}

The bootstrap's initial all-against-all leaderboard (\S{}4.1) is produced with the \textbf{same six criteria, output format, and JSON schema} as B.2, but with the calibration machinery removed. Relative to the anchored prompt above, the un-anchored form \textbf{omits} the following, and makes the substitutions noted:

\begin{enumerate}
\def\labelenumi{\arabic{enumi}.}
\tightlist
\item
  \textbf{Title line.} ``Score this submission against a calibration reference.'' becomes ``Score these. You are evaluating contest submissions.''
\item
  \textbf{The entire calibration preamble is removed} --- i.e.~everything from ``You are evaluating one contest submission ("Target Submission") against a fixed reference\ldots{}'' down to and including ``\ldots Do not score the Reference Submission itself --- its scores are fixed at \{ANCHOR\_\allowbreak{}SCORE\}.'' (the opening paragraph, the three ``Equal / Clearly better / Clearly worse'' bullets, and the ``Use the full 1--10 scale relative to the calibration anchor'' sentence).
\item
  \textbf{The scoring-instruction sentence drops its reference clause.} ``Score the Target Submission on six criteria, each rated 1--10 relative to the Reference (which is fixed at \{ANCHOR\_\allowbreak{}SCORE\} on every criterion)\ldots{} justifying the rating relative to the Reference'' becomes ``Score each submission on six criteria, each rated 1--10\ldots{} justifying the rating'' (all ``relative to the Reference'' qualifiers dropped).
\item
  \textbf{The Reference Submission block is removed.} The ``\#\# The submissions \ensuremath{\rightarrow} \#\#\# Reference Submission (fixed at \{ANCHOR\_\allowbreak{}SCORE\}/10\ldots) \{REFERENCE\_\allowbreak{}SUBMISSION\} \ensuremath{\rightarrow} \#\#\# Target Submission \ldots{} \{TARGET\_\allowbreak{}SUBMISSION\}'' section is replaced by a single batch: ``\#\# The proposals to evaluate'' followed by \texttt{\{SUBMISSIONS\}}.
\item
  \textbf{The output is per-submission, not per-target.} ``\#\# Target Submission'' becomes ``\#\# Submission '' repeated for each submission; all ``\textless\ldots relative to Reference\textgreater{}'' annotations in the output template are dropped; and the JSON top-level key changes from the single \texttt{"Target"} to one entry per \texttt{"\textless{}submission\_\allowbreak{}id\textgreater{}"}.
\item
  \textbf{The closing line drops its anchor clause.} ``All ratings are integers 1--10 inclusive. Equal to the Reference = \{ANCHOR\_\allowbreak{}SCORE\}.'' becomes ``All ratings are integers 1--10 inclusive.''
\end{enumerate}

Everything else --- the six criteria and their scope tags, the terminology block, the recursion note, and the per-archetype/per-PC/per-portfolio output structure --- is identical between the two forms.

\section{A council evaluation, worked}\label{a-council-evaluation-worked}

The released package contains one complete council evaluation from the canonical run: \textbf{gemini-2.5-flash}'s portfolio (the \emph{target}) --- a mid-leaderboard submission, strong enough to show the models play the Metanym Game competently yet flawed enough to draw substantive commentary --- scored by five evaluators: the four council seats other than the target, joined by claude-sonnet-4, the marginal case of \S{}4.3. The rubric operates at three levels and the record reproduces one unit of each in full: a \textbf{parallel context}, graded for factual truth sentence by sentence; an \textbf{archetype-level axis}, where the five score a whole archetype on one non-factual criterion; and the whole-portfolio structural-diversity judgement. Each unit shows the submitted material --- instantiation and idiomatic rewrite, metanyms capitalised --- with all five evaluators' ratings and comments and an administrator's synthesis of the anonymised council view (evaluators relabelled by a deterministic shuffle for the administrator's view only; the administrator is a Claude Opus model). Every evaluation in the run is released in the same form --- one file per (evaluator, portfolio) pair holding a factual score for each instantiated passage and a score on each of the other five axes, with justifications --- so any other unit can be inspected the same way. Three units are reproduced here, verbatim.

\subsection{A plain error, and the council converges}\label{a-plain-error-and-the-council-converges}

The first parallel context of the target's first archetype, \emph{Resource Allocation Under Scarcity}, instantiated in ecosystem management (Figure \ref{fig-game-evaluation} of the main text shows the same unit):

\begin{quote}
An ECOSYSTEM requires various NUTRIENTS to function and achieve its BIODIVERSITY. These NUTRIENTS are finite and often subject to PREDATION from other ECOSYSTEMS or internal GROWTH. NATURE must make NATURAL SELECTIONS about how to distribute the available NUTRIENTS among competing SPECIES. Misallocation of NUTRIENTS can lead to COLLAPSE of the ECOSYSTEM or hinder its ability to reach its BIODIVERSITY. Effective ECOLOGICAL SUCCESSION involves understanding the FOOD WEB of different NUTRIENTS and SPECIES, and adapting to changing CLIMATE. NATURE often faces an EXPLOITATION between short-term BIOMASS and long-term RESILIENCE.
\end{quote}

\textbf{claude-opus-4.5} --- Rating 5. ``The {[}instantiation{]} contains awkward phrasing like `Nature must make natural selections' which conflates natural selection (an evolutionary process) with resource allocation decisions. The statement `faces an exploitation between short-term biomass and long-term resilience' is grammatically incorrect and conceptually muddled. {[}idiomatic rewrite{]} improves clarity but still has issues---ecosystems don't really face `predation from other ecosystems.' The core concepts about nutrient cycling and ecological succession are roughly accurate but imprecisely stated compared to the Reference's precise biological terminology.''

\textbf{claude-opus-4.1} --- Rating 5. ``The parallel context contains basic factual accuracy but lacks the depth and precision of the Reference. The description of ecosystem resource allocation is overly simplified, treating `nature' as an allocator through natural selection, which is a conceptual stretch. The Reference's contexts demonstrate more sophisticated understanding of domain-specific mechanisms.''

\textbf{claude-opus-4.0} --- Rating 4. ``The parallel context contains several factual errors and awkward constructions. The phrase `predation from other ecosystems' misunderstands how ecosystems interact, and `nature must make natural selections' anthropomorphizes evolutionary processes incorrectly. The concept of ecosystems having `biodiversity' as a goal is teleological and scientifically inaccurate. Compared to the Reference's precise scientific language, this shows significantly weaker domain understanding.''

\textbf{claude-sonnet-4} --- Rating 4. ``The context contains several factual errors and awkward phrasings. `Nature must make natural selections' misrepresents natural selection as a conscious decision-making process, and `exploitation between short-term biomass and long-term resilience' uses `exploitation' incorrectly as a synonym for trade-off. {[}idiomatic rewrite{]} is more accurate but still contains the fundamental misunderstanding of natural selection as an active allocator. The Reference maintains better scientific accuracy throughout its parallel contexts.''

\textbf{gemini-3.1-pro} --- Rating 4. ``The mapping is somewhat strained. `Nature must make natural selections about how to distribute the available nutrients' is a teleological misrepresentation of natural selection, which is a consequence of differential survival, not an active allocator of nutrients. The Reference's mappings (e.g., bacterial chemotaxis) are much more precise and scientifically accurate.''

All five judges independently isolate the same clause, and the disagreement that remains --- 4 versus 5 --- is about severity, not about what is wrong. A submission cannot survive this by being persuasive; the claim either holds in the target domain or it does not.

\subsection{Judges disagreeing on a countable quantity}\label{judges-disagreeing-on-a-countable-quantity}

The same archetype scored on impressive length. Two of the five comments, juxtaposed:

\textbf{claude-opus-4.5} --- Rating 6. ``The template contains 15 slots and approximately 120 words, which is comparable to the Reference's templates. However, the template feels padded with generic concepts rather than building a complex, interconnected narrative. The Reference's templates of similar length achieve more conceptual density and structural sophistication.''

\textbf{gemini-3.1-pro} --- Rating 4. ``The context-template is 79 words long. The Reference's templates are generally longer (e.g., Gradient-Guided Navigation is 108 words, Containment Breach is 92 words). The Target's template is noticeably shorter and less detailed than the Reference.''

The object is a fixed string and the question is arithmetic, yet the two judges report 120 words and 79 words and reach opposite verdicts on the same criterion; across all five the reported counts range from 79 to 120. No judge can be treated as an oracle, and competence has to be estimated from the participants rather than assumed --- the motivation for \S{}3.3.

\subsection{The dissent that is better reasoned than the majority}\label{the-dissent-that-is-better-reasoned-than-the-majority}

The whole-portfolio structural-diversity judgement. Four evaluators rated 5--6; one rated 8. The majority view and the dissent:

\textbf{claude-opus-4.5} --- Rating 5. ``The six archetypes cover: (1) resource allocation, (2) information cascades, (3) homeostatic response, (4) emergence, (5) iterative refinement, and (6) decentralized coordination. These represent different system dynamics, but there is significant conceptual overlap---archetypes 4 and 6 both deal with emergence and self-organization, and ant colonies appear in both. The Reference's five archetypes (gradient navigation, containment breach, competitive exclusion, debt accumulation, scaffold assembly) show greater structural diversity, covering navigation, boundary dynamics, competition, temporal debt, and assembly processes. The Target's archetypes tend toward well-known systems concepts without the Reference's creative structural variety. The recursive claims in archetypes 4 and 5 are not convincingly demonstrated, unlike the Reference's elegant molecular\ensuremath{\rightarrow}civilizational hierarchy.''

\textbf{gemini-3.1-pro} --- Rating 8. ``The submitted set includes Resource Allocation, Information Cascade, Adaptive Response, Emergent Complexity, and Iterative Refinement. (Note: The submission actually contains six archetypes, but I am scoring the first five as per the standard format, though I will consider the sixth, `The Unseen Hand of Coordination,' in this diversity assessment as it was provided). These archetypes represent different system structures: allocation of finite resources, propagation of information/behavior, homeostatic regulation, bottom-up emergence, and goal-directed iteration. This is a very diverse set of system structures, arguably slightly more diverse than the Reference's set (which leans heavily on spatial/physical metaphors like navigation, containment, and scaffolding).''

The dissent is instructive rather than anomalous. gemini-3.1-pro advances a substantive counter-argument --- that the anchor's own set is biased toward spatial metaphors --- and is the only judge to notice and handle the fact that this portfolio contains six archetypes where the format specifies five. A rating that is both an outlier and better reasoned than the majority is the case a majority vote mishandles and a competence-weighted factorisation is meant to price (\S{}4.2).

\section{Auditing the metanym--GPQA correlation}\label{auditing-the-metanymgpqa-correlation}

\subsection{The correlation, decomposed}\label{the-correlation-decomposed}

\textbf{The aggregation ladder is monotone.} Council-basis official values against GPQA, three runs pooled, \(n = 12\) throughout:

\begin{table}[H]
\caption{The aggregation ladder. Two interval constructions are reported because each covers the other's weakness at \(n = 12\): Fisher-\(z\) assumes bivariate normality but unbends the skew of a bounded statistic; the BCa bootstrap is assumption-lighter and corrects the bias that makes the naive percentile bootstrap anti-conservative here. Where they disagree, the wider bound is the honest one.}
\begin{center}\small
\shrinktowidth{\begin{tabular}{@{}lrrll@{}}
\toprule\noalign{}
Quantity & Pearson \(r\) & Spearman \(\rho\) & Fisher-\(z\) 95\% & BCa bootstrap 95\% \\
\midrule\noalign{}
\(E^{C}\) alone &\cellcolor[HTML]{234A9E}\textcolor[HTML]{EEF3F8}{0.81}&\cellcolor[HTML]{24469D}\textcolor[HTML]{EEF3F8}{0.82}& {[}0.45, 0.95{]} & {[}0.42, 0.94{]}  \\
\(G^{F}\) alone &\cellcolor[HTML]{22318D}\textcolor[HTML]{EEF3F8}{0.89}&\cellcolor[HTML]{24409A}\textcolor[HTML]{EEF3F8}{0.84}& {[}0.64, 0.97{]} & {[}0.72, 0.95{]}  \\
\(E^{F}\) alone &\cellcolor[HTML]{253695}\textcolor[HTML]{EEF3F8}{0.87}&\cellcolor[HTML]{0F2366}\textcolor[HTML]{EEF3F8}{0.97}& {[}0.60, 0.96{]} & {[}0.76, 0.93{]}  \\
\(G^{C}\) alone &\cellcolor[HTML]{142670}\textcolor[HTML]{EEF3F8}{0.95}&\cellcolor[HTML]{24409A}\textcolor[HTML]{EEF3F8}{0.84}& {[}0.82, 0.99{]} & {[}0.82, 0.98{]}  \\
\(G = \tfrac12(G^{F}+G^{C})\) --- generation half &\cellcolor[HTML]{162875}\textcolor[HTML]{EEF3F8}{0.94}&\cellcolor[HTML]{253695}\textcolor[HTML]{EEF3F8}{0.87}& {[}0.81, 0.98{]} & {[}0.84, 0.98{]}  \\
\(E = \tfrac12(E^{F}+E^{C})\) --- evaluation half &\cellcolor[HTML]{162875}\textcolor[HTML]{EEF3F8}{0.94}&\cellcolor[HTML]{142670}\textcolor[HTML]{EEF3F8}{0.95}& {[}0.80, 0.98{]} & {[}0.83, 0.98{]}  \\
\(\tfrac12(E^{F}+G^{F})\) --- the factual pair &\cellcolor[HTML]{162875}\textcolor[HTML]{EEF3F8}{0.94}&\cellcolor[HTML]{11246B}\textcolor[HTML]{EEF3F8}{0.96}& {[}0.80, 0.98{]} & {[}0.89, 0.97{]}  \\
\(T = \tfrac14(G^{F}+G^{C}+E^{F}+E^{C})\) &\cellcolor[HTML]{0D2162}\textcolor[HTML]{EEF3F8}{0.98}&\cellcolor[HTML]{11246B}\textcolor[HTML]{EEF3F8}{0.96}& {[}0.93, 1.00{]} & {[}0.95, 0.99{]}  \\
\bottomrule
\end{tabular}}
\end{center}
\end{table}

\textbf{What each step adds.} The plain average of the ratings, every portfolio's leave-self-out mean over all axes and judges, tracks GPQA at \(r = 0.77\) un-anchored (the \S{}4.1 pass) and \(0.93\) anchored (run 1, the anchor at 7 by calibration); the total \(T\) reaches \(0.98\) (\texttt{scripts/baseline\_\allowbreak{}aggregators.py}).

Every component's interval sits well clear of zero --- GPQA corroborates \(T\) and, with varying strength, each component --- with \(T\)'s interval the tightest under both constructions. The four quarters are four \emph{differently distorted} reads of capability: \(G^{F}\) \textbf{ceilings at the top} (the leading eight compress into 6.23--7.00 while GPQA still spreads them across 19 points) and \textbf{offers a refuge at the bottom} (the GPT-4o family holds \(G^{F}\approx5.2\) on safe, simple, true portfolios while GPQA reads 46--48\% and \(G^{C}\) reads 3.5 --- truth rewards playing safe, beauty punishes it); \(E^{F}\) is judging in form but answering in content; \(E^{C}\) is a disposition that saturates once a model is competent enough to have a standard (gpt-4.1-mini's \(E^{C}\) of 6.97 sits above every Opus but the anchor). Equal-weight averaging cancels substantially independent distortions (Spearman--Brown); dropping even the weakest quarter lowers the aggregate (\(\tfrac13(G^{F}+G^{C}+E^{F})\) reads 0.97), and no sub-combination we examined beats the full average --- the best two-quarter pairing, \(\tfrac12(G^{C}+E^{F})\), reads 0.95. \(T\) is also the only compound with a pre-registered justification: it is the benchmark's official total, defined before any GPQA comparison, whereas any other weighting chosen for its GPQA agreement at \(n = 12\) would be curve-fitting.

\textbf{Measurement error, propagated.} Both coordinates carry known measurement distributions --- each model's \(T\) its replicate distribution, GPQA accuracy binomial --- so the fit can be re-derived with them propagated (slope point estimate 8.4 GPQA points per unit of \(T\)):

\begin{table}[H]
\caption{The \(T\)--GPQA fit with measurement uncertainty propagated. The last row is conservative (the observed scatter already contains one realisation of each point's noise); the correlation does not fall below 0.86. Measurement error in \(x\) attenuates a correlation rather than inflating it, so the point estimates are themselves conservative.}
\begin{center}\small
\begin{tabular}{@{}lll@{}}
\toprule\noalign{}
Uncertainty propagated & slope 95\% & \(r\) 95\% \\
\midrule\noalign{}
sampling of models only (pairs bootstrap) & {[}7.7, 9.7{]} & {[}0.96, 1.00{]} \\
measurement only (coordinate draws, models fixed) & {[}6.9, 9.7{]} & {[}0.90, 0.98{]} \\
both simultaneously & {[}6.6, 10.3{]} & {[}0.86, 0.98{]} \\
\bottomrule
\end{tabular}
\end{center}
\end{table}

\textbf{The regimes invert --- \(T\) is the only regime-invariant indicator.} Restricting to the leading eight reverses the single-quarter ordering:

\begin{table}[H]
\caption{Pearson \(r\) against GPQA on the full roster and on the leading eight (point estimates on eight points).}
\begin{center}\small
\begin{tabular}{@{}lrr@{}}
\toprule\noalign{}
Quantity & Full roster (\(n=12\)) & Leading eight (\(n=8\)) \\
\midrule\noalign{}
\(G^{F}\) &\cellcolor[HTML]{22318D}\textcolor[HTML]{EEF3F8}{0.89}&\cellcolor[HTML]{1F7FB7}\textcolor[HTML]{EEF3F8}{0.67} \\
\(G^{C}\) &\cellcolor[HTML]{142670}\textcolor[HTML]{EEF3F8}{0.95}&\cellcolor[HTML]{234A9E}\textcolor[HTML]{EEF3F8}{0.81} \\
\(E^{F}\) &\cellcolor[HTML]{253695}\textcolor[HTML]{EEF3F8}{0.87}&\cellcolor[HTML]{22318D}\textcolor[HTML]{EEF3F8}{0.89} \\
\(E^{C}\) &\cellcolor[HTML]{234A9E}\textcolor[HTML]{EEF3F8}{0.81}&\cellcolor[HTML]{82CEBB}\textcolor[HTML]{1A1A1A}{0.37} \\
\(\tfrac12(E^{F}+G^{F})\) &\cellcolor[HTML]{162875}\textcolor[HTML]{EEF3F8}{0.94}&\cellcolor[HTML]{1B2C7E}\textcolor[HTML]{EEF3F8}{0.92} \\
\(\tfrac12(G^{C}+E^{F})\) &\cellcolor[HTML]{142670}\textcolor[HTML]{EEF3F8}{0.95}&\cellcolor[HTML]{162875}\textcolor[HTML]{EEF3F8}{0.94} \\
\(T\) &\cellcolor[HTML]{0D2162}\textcolor[HTML]{EEF3F8}{0.98}&\cellcolor[HTML]{162875}\textcolor[HTML]{EEF3F8}{0.94} \\
\bottomrule
\end{tabular}
\end{center}
\end{table}

Across the full roster the capability cliff gives the subjective making axis its discriminating range; among the elite every model is a competent maker (gemini-3.1-pro tops GPQA yet is a middling maker), and what still separates frontier models is knowledge, which is \(E^{F}\)'s content: detecting errors in others' work stays hard after producing clean work has become easy, so \(E^{F}\) keeps its spread (7.78 down to 0.77 across the eight) exactly where \(G^{F}\) compresses. Within the Anthropic family alone (\(n = 4\)) the rank agreement is \(\rho = 0.80\) --- where the evaluators leave the ordering unresolved, the agreement with GPQA coarsens too. Recomputing every quantity on the twelve-evaluator basis instead of the contest basis moves \(T\)'s correlation from 0.982 to 0.976; the one real difference is \(E^{C}\) (0.81 contest vs 0.89 twelve-evaluator), the contest's easier consistency test inflating inert-band judges in a way an external instrument can see. Taken apart, the three runs read \(r = 0.97, 0.97, 0.92\) (\(\rho = 0.93, 0.95, 0.88\)), the dip being run 3, whose factual axis is unidentified (\(\sigma_1/\sigma_2 = 1.29\)); pooled, 0.98 (\(\rho = 0.96\)); excluding the anchor changes nothing (0.97, 0.97, 0.91; pooled 0.98).

\textbf{What it does and does not corroborate.} \(T\) and GPQA are both broad capability measures, so their agreement corroborates the benchmark as a whole. It does not by itself certify that the factual axis recovered \emph{truth} rather than capability-correlated quality: with \(G^{C}\) alone at 0.95, GPQA concordance is not an axis-specific claim.

\subsection{The administration, disclosed}\label{the-administration-disclosed}

GPQA Diamond (198 questions) was administered 2026-06-13 --- two weeks \emph{after} the generation run (2026-05-29), so no direction exists for the benchmark's content to have been shaped by GPQA. All twelve models were queried through the same gateway and protocol as the council run (Temperature 0, no dedicated reasoning channel, no tools), with the question and four shuffled options only --- no key ever enters a prompt:

\begin{lstlisting}
Answer the following multiple choice question. The last line of your reply
must be exactly 'Answer: $LETTER' where $LETTER is one of A, B, C, D.

{question}

A) {A}
B) {B}
C) {C}
D) {D}
\end{lstlisting}

Option order is shuffled deterministically per question (seeded by question index, identically for every model).

\textbf{The administration was two-stage.} ``No dedicated reasoning channel'' is not deliberation-off: models write visible derivations of vendor-idiosyncratic length before the answer line, and under the initial 2,048-token cap the two Gemini seats were massively truncated --- the first pass (preserved in the run's log) scored gemini-3.1-pro at 82/198 with 103 unparseable responses and gemini-2.5-flash at 121/198 with 50, voids counted as wrong. The cap was raised to 8,192 and the void responses --- only the void ones --- were re-asked; the published records are the patched set (13 residual voids per Gemini seat, still counted as wrong). Two facts bound the bias: retries targeted only \emph{unparseable} responses, never parsed-but-wrong answers; and the retry success rate did not exceed the first pass's scored-only accuracy on either seat (gemini-3.1-pro: 78/90 retried items correct, 86.7\%, vs 86.3\% on its 95 parseable first-pass responses --- above the published 80.81\%; gemini-2.5-flash: 22/37, 59\%, vs 81.8\%). Both stages' evidence ships with the package.

\subsection{The audit: no leak found}\label{the-audit-no-leak-found}

An audit script re-derives the published table from the raw records and hard-fails on any mismatch. Its checks, all passing:

\begin{enumerate}
\def\labelenumi{\arabic{enumi}.}
\tightlist
\item
  \textbf{CSV reconciliation} --- per-model correct counts recomputed from raw equal the published accuracies exactly, all twelve models.
\item
  \textbf{Key consistency} --- the key letter is identical across all twelve models' records for every question.
\item
  \textbf{Key balance} --- shuffled key distribution A/B/C/D = 48/51/47/52, chi-square vs uniform \(p = 0.95\).
\item
  \textbf{Independent re-extraction} --- a second answer extractor, written without sight of the first, agrees with the stored verdicts on every response for ten of twelve models; the Gemini disagreements are almost entirely the independent extractor failing on the LaTeX \texttt{\textbackslash{}boxed\{X\}} answer style. Genuinely questionable credits --- granted with no terminal answer statement --- number 2 (gemini-2.5-flash) and 6 (gemini-3.1-pro) of 198.
\item
  \textbf{Strict-terminal sensitivity} --- rescoring with only explicit terminal answer statements counted moves three models (gemini-2.5-flash 72.22 \ensuremath{\rightarrow} 70.71, gemini-3.1-pro 80.81 \ensuremath{\rightarrow} 77.27, gpt-4.1-mini 63.13 \ensuremath{\rightarrow} 60.10) and the \(T\)--GPQA correlation from 0.982 to 0.972. Scored the opposite way --- voids excluded rather than counted wrong --- it reads 0.975.
\item
  \textbf{First-pass log reconciliation} --- the archived first-pass log reconciles exactly with the shipped two-stage records.
\end{enumerate}

\textbf{What the audit cannot rule out.} It certifies the path from raw records to published numbers; it cannot re-run the models. Both instruments share the gateway, so capability-correlated properties of that path touch both, and the gateway's \texttt{thinkingBudget:\ 0} for the Gemini seats is passed but not verified by assertion. And GPQA is public: \emph{uniform} training contamination would inflate accuracies without inflating a correlation against freshly generated items, but \emph{differential} contamination --- exposure increasing with training recency and scale, which correlate with capability --- would inflate the slope itself. That channel cannot be excluded with the shipped data and is the specific residual threat.

\begin{table}[H]
\caption{The two instruments side by side --- the key-free total \(T\) (three runs pooled) and self-administered GPQA Diamond accuracy (voids counted as wrong), sorted by \(T\).}
\label{tab-gpqa-values}
\begin{center}\small
\begin{tabular}{@{}lrr@{}}
\toprule\noalign{}
Model & \(T\) & GPQA Diamond (\%) \\
\midrule\noalign{}
\ensuremath{\star} claude-opus-4.5 (anchor) &\cellcolor[HTML]{2072B2}\textcolor[HTML]{EEF3F8}{7.00}&\cellcolor[HTML]{2351A2}\textcolor[HTML]{EEF3F8}{78.79} \\
gemini-3.1-pro &\cellcolor[HTML]{2076B3}\textcolor[HTML]{EEF3F8}{6.92}&\cellcolor[HTML]{234A9F}\textcolor[HTML]{EEF3F8}{80.81} \\
claude-opus-4.1 &\cellcolor[HTML]{2498C1}\textcolor[HTML]{EEF3F8}{6.02}&\cellcolor[HTML]{2258A5}\textcolor[HTML]{EEF3F8}{76.77} \\
claude-opus-4.0 &\cellcolor[HTML]{2A9EC1}\textcolor[HTML]{EEF3F8}{5.81}&\cellcolor[HTML]{206DAF}\textcolor[HTML]{EEF3F8}{71.21} \\
gemini-2.5-flash &\cellcolor[HTML]{2BA0C2}\textcolor[HTML]{EEF3F8}{5.75}&\cellcolor[HTML]{2169AD}\textcolor[HTML]{EEF3F8}{72.22} \\
claude-sonnet-4 &\cellcolor[HTML]{37ACC3}\textcolor[HTML]{EEF3F8}{5.34}&\cellcolor[HTML]{2169AD}\textcolor[HTML]{EEF3F8}{72.22} \\
gpt-4.1-mini &\cellcolor[HTML]{40B5C4}\textcolor[HTML]{1A1A1A}{5.03}&\cellcolor[HTML]{1D8EBF}\textcolor[HTML]{EEF3F8}{63.13} \\
gpt-4.1-2025-04-14 &\cellcolor[HTML]{52BCC2}\textcolor[HTML]{1A1A1A}{4.66}&\cellcolor[HTML]{2094C0}\textcolor[HTML]{EEF3F8}{61.62} \\
gpt-4.1-nano &\cellcolor[HTML]{83CFBB}\textcolor[HTML]{1A1A1A}{3.68}&\cellcolor[HTML]{32A7C2}\textcolor[HTML]{EEF3F8}{55.05} \\
gpt-4o-2024-08-06 &\cellcolor[HTML]{97D6B9}\textcolor[HTML]{1A1A1A}{3.34}&\cellcolor[HTML]{53BDC1}\textcolor[HTML]{1A1A1A}{46.46} \\
gpt-4o &\cellcolor[HTML]{ADDFB7}\textcolor[HTML]{1A1A1A}{2.95}&\cellcolor[HTML]{49B9C3}\textcolor[HTML]{1A1A1A}{48.48} \\
gpt-4o-mini &\cellcolor[HTML]{CAEAB4}\textcolor[HTML]{1A1A1A}{2.39}&\cellcolor[HTML]{5FC1C0}\textcolor[HTML]{1A1A1A}{43.94} \\
\bottomrule
\end{tabular}
\end{center}
\end{table}

\section{The constructs of intelligence the Metanym Game demands}\label{the-constructs-of-intelligence-the-metanym-game-demands}

Playing the metanym game demands at least eight constructs that cognitive science treats as central to intelligence, abstraction and analogy above all --- one wide cluster, not the whole of it (perception, motor skill, working memory and social intelligence the Metanym Game leaves alone):

\begin{enumerate}
\def\labelenumi{\arabic{enumi}.}
\tightlist
\item
  \textbf{Higher-order relational reasoning} (Penn et al., 2008) --- recognising that two situations share the same pattern of relations among their parts even when the parts are unrelated. Each slot is defined by its relations to the other slots, not by its filler word; successful substitution shows the pattern survives.
\item
  \textbf{Structure-mapping} (Gentner, 1983; Falkenhainer et al., 1989) --- the metanym table is the structure-mapping bookkeeping written down: mechanical substitutability enforces one-to-one correspondence and parallel connectivity, while systematicity is judged rather than assumed (the \texttt{intelligence} axis).
\item
  \textbf{Essence-seeing} (Hofstadter \& Sander, 2013) --- spotting that a novel situation is an instance of a known abstract pattern: seeing the archetypal context behind the template.
\item
  \textbf{Fluid intelligence} (Cattell, 1963; Horn \& Cattell, 1966) --- reasoning out a novel structure on the fly.
\item
  \textbf{Crystallised intelligence} (Cattell, 1963; Horn \& Cattell, 1966) --- knowing the vocabulary and domain facts that make every substituted sentence true.
\item
  \textbf{Convergent production} (Guilford, 1967) --- substitution admits no near-misses; a sentence passes or fails.
\item
  \textbf{Divergent production} (Guilford, 1967) --- the same template must travel to widely different domains, each with its own metanym set.
\item
  \textbf{Theory formation by analogy} (Hesse, 1963) --- each archetypal context is theory construction in miniature: the template is the root analogy, each parallel context extends it into new territory, and factuality is the empirical test.
\end{enumerate}

\begin{table}[H]
\caption{The eight constructs mapped to the Metanym Game's two tasks (\ensuremath{\bullet} marks a primary demand).}
\label{tab-constructs}
\begin{center}\small
\shrinktowidth{\begin{tabular}{@{}lll@{}}
\toprule\noalign{}
Construct & Generation --- invent the template & Evaluation --- judge a portfolio \\
\midrule\noalign{}
Higher-order relational reasoning & \ensuremath{\bullet} lay out the relational skeleton & \ensuremath{\bullet} check the relations survive \\
Structure-mapping & \ensuremath{\bullet} the slots-and-domains scaffold & \ensuremath{\bullet} verify one-to-one correspondence \\
Essence-seeing & \ensuremath{\bullet} see the archetype behind the surface & \\
Fluid intelligence & \ensuremath{\bullet} reason out a novel structure & \\
Crystallised intelligence & \ensuremath{\bullet} supply true domain terms and facts & \ensuremath{\bullet} detect false claims \\
Convergent production & & \ensuremath{\bullet} pass/fail, no near-misses \\
Divergent production & \ensuremath{\bullet} invent a structure that travels & \\
Theory formation by analogy & \ensuremath{\bullet} template = root analogy, tested by fact & \\
\bottomrule
\end{tabular}}
\end{center}
\end{table}

Breadth matters because general ability is, empirically, what broad batteries measure --- across 591 language models and twelve diverse tests, a single general factor explains about two thirds of performance variance (Ili\'{c} \& Gignac, 2024). The total rating's agreement with GPQA (\S{}4.5, \S{}6) is consistent with that, and the paper offers it as no more. Intelligence has no ground truth and never did: psychology's own definition is an aggregate of individual conceptions --- Neisser (1979) argued that ``intelligent'' is a prototype concept assembled from people's judgements, and Sternberg et al.~(1981) measured people's conceptions of it by pooling personal ratings --- and objectivity, in the philosophy of science, is constituted by a community holding individual judgements to mutual criticism under shared standards (Longino, 1990). The benchmark mechanises exactly this collective--individual link: individually, a judge must hold a firm standard, stable as the tare shifts; collectively, trust flows to the judge the other trusted judges agree with.

\section{The three runs apart, and pooled}\label{the-three-runs-apart-and-pooled}

\begin{table}[H]
\caption{Total rating \(T\) across three full re-runs (run 1 = the bootstrap generation, re-analysed on the council basis; runs 2--3 the same day, two hours apart), all on the council basis. \ensuremath{\star} marks a council seat, fixed from run 1. SD is the per-model run-to-run standard deviation (mean 0.43, max 0.72). Pairwise agreement between runs: Pearson 0.92--0.96, Spearman 0.84--0.90. Re-selected from a single run's totals, the council would seat gpt-4.1-mini for claude-opus-4.0 in run 2 and claude-sonnet-4 for gemini-2.5-flash in run 3; neither rotation clears the guard (contest \(\sigma_1/\sigma_2\) 1.87 and 1.63, \protect\texttt{scripts/per\_\allowbreak{}run\_\allowbreak{}contests.py}). Run 3's \(E^{F}\) quarter is read off a factual axis the guard reports as unidentified (\(\sigma_1/\sigma_2 = 1.29\)).}
\label{tab-reruns}
\begin{center}\small
\begin{tabular}{@{}lrrrr@{}}
\toprule\noalign{}
Model & \(T_1\) & \(T_2\) & \(T_3\) & SD \\
\midrule\noalign{}
\ensuremath{\star} claude-opus-4.5 &\cellcolor[HTML]{2072B2}\textcolor[HTML]{EEF3F8}{7.00}&\cellcolor[HTML]{2072B2}\textcolor[HTML]{EEF3F8}{7.00}&\cellcolor[HTML]{2072B2}\textcolor[HTML]{EEF3F8}{7.00}&\cellcolor[HTML]{FFFFD9}\textcolor[HTML]{1A1A1A}{0.00} \\
\ensuremath{\star} gemini-3.1-pro &\cellcolor[HTML]{1F81B8}\textcolor[HTML]{EEF3F8}{6.65}&\cellcolor[HTML]{2164AB}\textcolor[HTML]{EEF3F8}{7.35}&\cellcolor[HTML]{234A9E}\textcolor[HTML]{EEF3F8}{8.10}&\cellcolor[HTML]{216AAE}\textcolor[HTML]{EEF3F8}{0.72} \\
\ensuremath{\star} claude-opus-4.0 &\cellcolor[HTML]{2397C1}\textcolor[HTML]{EEF3F8}{6.04}&\cellcolor[HTML]{33A7C2}\textcolor[HTML]{EEF3F8}{5.50}&\cellcolor[HTML]{1D91C0}\textcolor[HTML]{EEF3F8}{6.26}&\cellcolor[HTML]{78CABC}\textcolor[HTML]{1A1A1A}{0.39} \\
\ensuremath{\star} gemini-2.5-flash &\cellcolor[HTML]{2398C1}\textcolor[HTML]{EEF3F8}{6.03}&\cellcolor[HTML]{299DC1}\textcolor[HTML]{EEF3F8}{5.84}&\cellcolor[HTML]{4CBAC2}\textcolor[HTML]{1A1A1A}{4.78}&\cellcolor[HTML]{1F7BB5}\textcolor[HTML]{EEF3F8}{0.68} \\
\ensuremath{\star} claude-opus-4.1 &\cellcolor[HTML]{279BC1}\textcolor[HTML]{EEF3F8}{5.92}&\cellcolor[HTML]{1F7CB6}\textcolor[HTML]{EEF3F8}{6.77}&\cellcolor[HTML]{1F93C0}\textcolor[HTML]{EEF3F8}{6.19}&\cellcolor[HTML]{64C3BF}\textcolor[HTML]{1A1A1A}{0.43} \\
claude-sonnet-4 &\cellcolor[HTML]{3BAFC3}\textcolor[HTML]{1A1A1A}{5.22}&\cellcolor[HTML]{299EC1}\textcolor[HTML]{EEF3F8}{5.82}&\cellcolor[HTML]{2094C0}\textcolor[HTML]{EEF3F8}{6.16}&\cellcolor[HTML]{50BCC2}\textcolor[HTML]{1A1A1A}{0.47} \\
gpt-4.1-mini &\cellcolor[HTML]{47B8C3}\textcolor[HTML]{1A1A1A}{4.88}&\cellcolor[HTML]{269AC1}\textcolor[HTML]{EEF3F8}{5.93}&\cellcolor[HTML]{44B7C3}\textcolor[HTML]{1A1A1A}{4.93}&\cellcolor[HTML]{279BC1}\textcolor[HTML]{EEF3F8}{0.59} \\
gpt-4.1-2025-04-14 &\cellcolor[HTML]{5DC0C0}\textcolor[HTML]{1A1A1A}{4.44}&\cellcolor[HTML]{54BDC1}\textcolor[HTML]{1A1A1A}{4.62}&\cellcolor[HTML]{2CA1C2}\textcolor[HTML]{EEF3F8}{5.72}&\cellcolor[HTML]{2076B4}\textcolor[HTML]{EEF3F8}{0.69} \\
gpt-4.1-nano &\cellcolor[HTML]{92D4B9}\textcolor[HTML]{1A1A1A}{3.42}&\cellcolor[HTML]{8BD1BA}\textcolor[HTML]{1A1A1A}{3.55}&\cellcolor[HTML]{BAE4B5}\textcolor[HTML]{1A1A1A}{2.72}&\cellcolor[HTML]{5ABFC0}\textcolor[HTML]{1A1A1A}{0.45} \\
gpt-4o-2024-08-06 &\cellcolor[HTML]{97D6B9}\textcolor[HTML]{1A1A1A}{3.34}&\cellcolor[HTML]{A6DCB7}\textcolor[HTML]{1A1A1A}{3.08}&\cellcolor[HTML]{86D0BA}\textcolor[HTML]{1A1A1A}{3.63}&\cellcolor[HTML]{BBE5B5}\textcolor[HTML]{1A1A1A}{0.27} \\
gpt-4o &\cellcolor[HTML]{B3E1B6}\textcolor[HTML]{1A1A1A}{2.85}&\cellcolor[HTML]{B8E3B5}\textcolor[HTML]{1A1A1A}{2.76}&\cellcolor[HTML]{AEDFB6}\textcolor[HTML]{1A1A1A}{2.93}&\cellcolor[HTML]{F2FABC}\textcolor[HTML]{1A1A1A}{0.09} \\
gpt-4o-mini &\cellcolor[HTML]{D3EEB3}\textcolor[HTML]{1A1A1A}{2.10}&\cellcolor[HTML]{CFECB3}\textcolor[HTML]{1A1A1A}{2.25}&\cellcolor[HTML]{E2F4B2}\textcolor[HTML]{1A1A1A}{1.60}&\cellcolor[HTML]{93D5B9}\textcolor[HTML]{1A1A1A}{0.34} \\
\bottomrule
\end{tabular}
\end{center}
\end{table}

The factual quarter is where the runs differ. On the twelve-evaluator basis, with the consistency ratings from run 1's sweep throughout:

\begin{table}[H]
\caption{Factual competence \(E^{F}\) per run and pooled (twelve-evaluator basis). Spectral gap \(\sigma_1/\sigma_2\): 2.57, 1.67, 1.29; pooled 2.24 (runs 1 and 2 alone, 2.45). Run-to-run agreement (Pearson) of \(E^{F}\) alone: 0.90, 0.79, 0.80 for runs 1--2, 1--3, 2--3; of the three regenerated quarters \(\tfrac13(G^{F}+G^{C}+E^{F})\): 0.95, 0.91, 0.93; of \(T\): 0.97, 0.94, 0.95.}
\label{tab-ef-runs}
\begin{center}\small
\begin{tabular}{@{}lrrrr@{}}
\toprule\noalign{}
Model & \(E^{F}\) run 1 & run 2 & run 3 & pooled \\
\midrule\noalign{}
gemini-3.1-pro & 7.36 & 8.29 & 10.24 & 8.24 \\
\ensuremath{\star} claude-opus-4.5 & 7.00 & 7.00 & 7.00 & 7.00 \\
claude-opus-4.0 & 4.47 & 3.48 & 5.41 & 4.69 \\
claude-opus-4.1 & 3.55 & 6.78 & 5.17 & 4.65 \\
gemini-2.5-flash & 4.68 & 3.81 & 0.01 & 3.47 \\
claude-sonnet-4 & 1.62 & 2.44 & 4.00 & 2.37 \\
gpt-4.1-mini & 1.21 & 3.88 & 0.06 & 1.34 \\
gpt-4.1-2025-04-14 & 0.20 & 0.53 & 2.92 & 0.79 \\
gpt-4o & 0.34 & 0.41 & 1.02 & 0.47 \\
gpt-4o-2024-08-06 & 0.62 & 0.00 & 0.00 & 0.35 \\
gpt-4.1-nano & 0.49 & 0.39 & 0.00 & 0.28 \\
gpt-4o-mini & 0.00 & 0.00 & 0.00 & 0.00 \\
\bottomrule
\end{tabular}
\end{center}
\end{table}

\begin{figure}[H]
\centering
\pandocbounded{\includegraphics[width=1.00\linewidth]{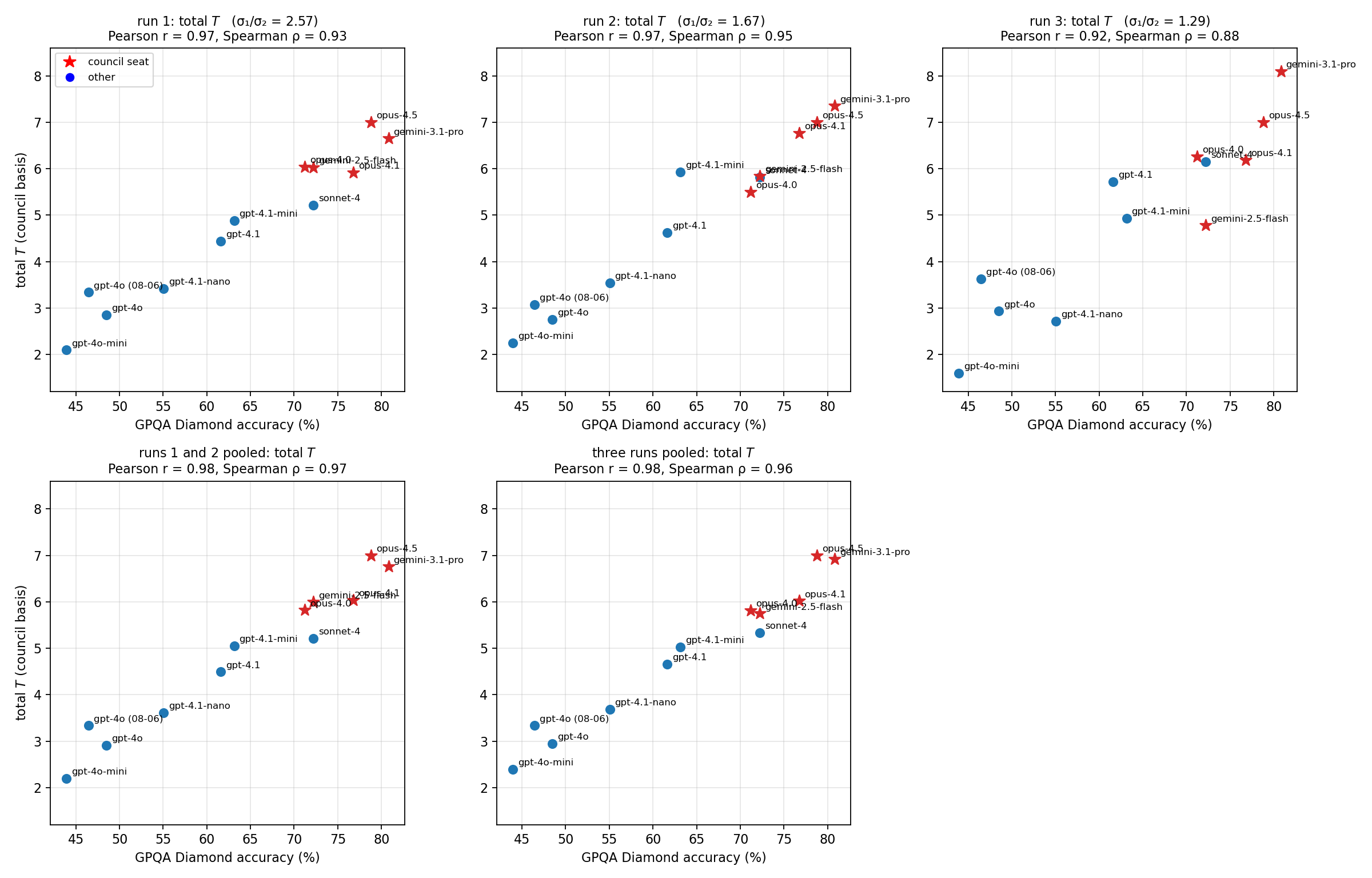}}
\caption{The three runs apart and pooled, against GPQA Diamond. Top row: council-basis \(T\) per run, with each run's twelve-evaluator spectral gap. Bottom row: runs 1 and 2 pooled (left), all three pooled (right). Run 3, with the smallest gap, is the odd one out; pooling restores an identified factual axis and the agreement rises.}
\end{figure}

\(\sigma_1/\sigma_2\) is the standard spectral-gap diagnostic for whether a leading singular direction is identified (Davis--Kahan: the rotation of \(u_1\) under perturbation is bounded by the gap to \(\sigma_2\) and by nothing below it); the threshold of 2 is a rule of thumb, not a derived value. Run 2 sits below it and reads well; run 3 sits further below and does not. On this evidence a threshold near 1.5 may separate the two, but two runs cannot fix it, and GPQA, the external reference that names run 3 as the odd one out, is not ground truth for intelligence and sets no protocol parameter here. The benchmark's policy is therefore to pool: three portfolios per player, the seats' and the ballast's three stored, one factorisation over all of them, and the guard reported on the pooled matrix.

\section{What anchoring does to resolution}\label{what-anchoring-does-to-resolution}

\begin{figure}[H]
\centering
\pandocbounded{\includegraphics[width=0.60\linewidth]{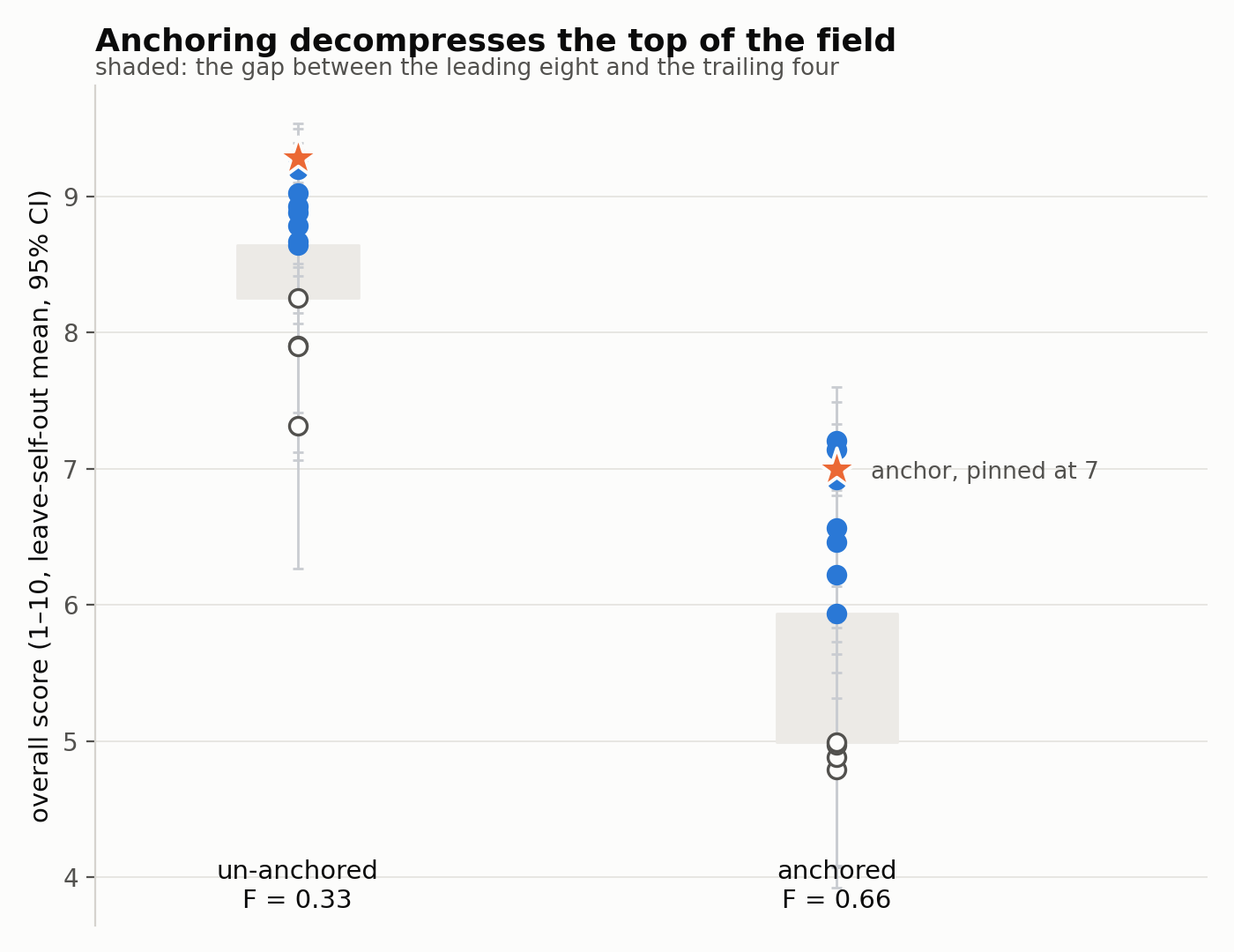}}
\caption{Each model's leave-self-out overall mean with its 95\% bootstrap CI, un-anchored (left) and anchored (right); models are unlabelled deliberately --- the message is structural, not a ranking. Filled markers are the leading eight, open the trailing four, the shaded band the gap between them; the star is the anchor --- a scored target on the left, the pinned reference (7) on the right. Anchored scores spread around the yardstick instead of piling against the ceiling: the gap more than doubles relative to the spread of the means (every scored cross-band pair holds at bootstrap probability 1.00), the variance ratio --- between-target over within-target variance, Fisher's \(F\) --- doubles from 0.33 to 0.66, and the intervals still overlap within each band.}
\label{fig-anchoring-resolution}
\end{figure}

\end{document}